\documentclass[twoside]{article}
\pdfoutput=1

\usepackage[a4paper]{geometry}
\usepackage[latin1]{inputenc} 
\usepackage[T1]{fontenc} 
\usepackage{RR}
\usepackage{hyperref}
\usepackage{graphics} 
\usepackage{graphicx}
\usepackage{epsfig} 
\usepackage{amsmath} 
\usepackage{amssymb}  
\usepackage{latexsym}

\usepackage{bm}             
\usepackage{epstopdf}

\usepackage{algorithm}
\usepackage{algorithmic}

\usepackage{array}
\usepackage{mdwmath}
\usepackage{mdwtab}
\usepackage{multirow}
\usepackage{url}
\usepackage{color,soul}
\usepackage{subcaption}

\graphicspath{{figures/}}

\RRNo{8767}

\RRdate{August 2015}

\RRauthor{
K. Avrachenkov\thanks{K. Avrachenkov is with Inria Sophia Antipolis, France,
{\tt\small k.avrachenkov@inria.fr}}%
  \and
V.S. Borkar \thanks{V.S. Borkar is with Department of Electrical Engineering, IIT Bombay, India,
{\tt\small borkar.vs@gmail.com }}%
  \and
K. Saboo\thanks{K. Saboo is with Department of Electrical Engineering, IIT Bombay, India,
{\tt\small kvsaboo.2004@ee.iitb.ac.in}}
\thanks[sfn]{The work of KA and VSB was supported in part by grant no.\ 5100-ITA from the
Indo-French Centre for the Promotion of Advanced Research (IFCPAR) and
Alcatel-Lucent Inria Joint Lab.}
}
\authorhead{K. Avrachenkov \& V.S. Borkar \& K. Saboo}

\title{Parallel and Distributed Approaches for\\ Graph Based Semi-supervised Learning}

\RRtitle{Les Approches Parall\`eles et Distribu\'es pour l'Apprentissage Semi-supervis\'e}
\RRetitle{Parallel and Distributed Approaches for\\ Graph Based Semi-supervised Learning}
\titlehead{Distributed Semi-supervised Learning}

\RRresume{
Deux approches pour l'apprentissage semi-supervis\'e bas\'e sur le graphe de similarit\'e
sont propos\'es.
La premi\`ere approche est bas\'ee sur l'it\'eration d'un op\'erateur affine. Un \'el\'ement cl\'e de
l'it\'eration de l'op\'erateur affine est la multiplication vecteur par matrice de type sparse,
qui a plusieurs impl\'ementations parall\`eles tr\`es efficaces.
La seconde approche appartient \`a la classe des algorithmes de Monte-Carlo par cha\^ines de Markov (MCMC).
Elle est bas\'e sur un \'echantillonnage de noeuds en effectuant une marche al\'eatoire sur le graphe
de similarit\'e. Cette derni\`ere approche est distribu\'e par sa nature et peut \^etre
facilement mis en oeuvre sur plusieurs processeurs ou sur un r\'eseau.
\'Evaluations th\'eoriques ainsi que pratiques sont fournis.
On constate que les noeuds sont class\'es dans leurs classes avec tr\`es petite erreur.
La capacit\'e de l'algorithme MCMC de suivre les nouveaux noeuds arrivants online
et de les classer est \'egalement d\'emontr\'e.
}
\RRabstract{
Two approaches for graph based semi-supervised learning are proposed.
The first approach is based on iteration of an affine map.
A key element of the affine map iteration is sparse matrix-vector multiplication,
which has several very efficient parallel implementations. The second approach belongs
to the class of Markov Chain Monte Carlo (MCMC) algorithms.
It is based on sampling of nodes by performing a random walk on the graph.
The latter approach is distributed by its nature and can be
easily implemented on several processors or over the network.
Both theoretical and practical evaluations are provided.
It is found that the nodes are classified into their class with very small error.
The sampling algorithm's ability to track new incoming nodes and to classify them
is also demonstrated.
}

\RRmotcle{Apprentissage semi-supervis\'e, Apprentissage bas\'e sur le graphe de similarit\'e,
Algorithmes distribu\'es, Algorithmes parall\`eles}
\RRkeyword{Semi-supervised learning, Graph-based learning, Distributed algorithms, Parallel algorithms}

\RRprojet{Maestro}

\RCSophia

\begin{document}

\makeRR

\tableofcontents
\newpage

\section{Introduction}

Semi-supervised learning is a special type of learning which uses labelled as well as unlabelled data for
training \cite{CSZ06}.
In reality, the amount of labelled data is much less compared to the unlabelled data.
This makes this type of learning a powerful tool for processing the huge amount of data available nowadays.
The present work focuses on graph based semi-supervised learning \cite{CSZ06,FoussPirotte06Ker,FoussFrancoisseSaerens11,Z05}. Consider a (weighted) graph in which nodes belong to one of $K$ classes and the true class of a few nodes is given.
An edge and its weight of the graph indicate a similarity and similarity degree, respectively, between two nodes.
Thus, such graph is called the similarity graph. We assume that the (weighted) similarity graph is known.
Semi-supervised learning aims at using the true class information of few labelled nodes and the structure of the graph
to estimate the class of each of the remaining nodes.

Graph based semi-supervised learning finds application in recommendation systems \cite{FoussPirotte06Ker},
classification of web pages into categories \cite{ZHS04}, person name disambiguation \cite{SAT10}, etc.
Being of practical importance, graph based semi-supervised learning has been widely studied. Most methods formulate the learning as an optimization problem for feature vectors and then solve it iteratively or by deriving closed form solutions.
The problem assumes smoothness of node features across the graph and penalizes the deviation of a feature from
the true class of a node in case the true class is known.
Iterative solutions become computationally intensive because they involve matrix operations which grow in size as a polynomial with the graph size, though the growth can be linear or quasi-linear in the case of sparse graphs. In \cite{semisup} the authors have formulated a generalized optimization problem which gives as important particular cases:
Standard Laplacian \cite{ZB07}, Normalised Laplacian \cite{Zetal04} and PageRank \cite{Aetal08} algorithms.

In the present work, starting from the generalized formulation of \cite{semisup}, we propose two approaches for the
computation of the feature vectors. The first approach is based on iteration of an affine map. A key element of the affine map iteration is sparse matrix-vector multiplication, which has several very efficient parallel implementations.

The second approach belongs to the class of Markov Chain Monte Carlo algorithms.
The concept of sampling as in reinforcement learning is useful because it provides room for developing distributed computing schemes which reduce cost of per iterate computation, a crucial advantage in large problems. The realm of reinforcement learning is similar to semi-supervised learning in that the labelled data is suggestive but not exact. Borkar et al. \cite{reinf} have developed a sampling based reinforcement learning scheme for the PageRank algorithm. At each step of the iteration, a node is sampled uniformly from all the nodes. A neighbour of the sampled node is chosen according to the conditional probability of the neighbour given the sampled node. The stationary probability of the sampled node is then updated using a stochastic approximation of the ODE involving stationary distribution and the PageRank matrix.
This approach can be implemented in a distributed setting and hence the requirement of the entire adjacency matrix being known locally can also be relaxed, since information regarding only a neighbourhood is required.

The rest of the article is organised as follows: Section \ref{sec2} gives the algorithms, Section \ref{sec3} proves their convergence, Section \ref{sec4} details the experimental results and Section \ref{sec5} concludes with some observations.

\section{Optimization Problem and Algorithms}
\label{sec2}

Consider a graph with adjacency matrix $A$ which has $N$ nodes, each one belonging to one of $K$ classes. The class information of some of the nodes is also given, which will be referred to as labelled nodes. Define $D$ as a diagonal matrix with $D_{ii} = d(i)$, where $d(i)$ is the degree of node $i$. $Y$ is a $N \times K$ matrix which contains the information about the labelled nodes. More specifically,

\[ Y_{ij} = \left\{
  \begin{array}{l l}
    1 & \quad \text{if labelled node $i$ belongs to class $j$,}\\
    0 & \quad \text{otherwise.}
  \end{array} \right.\]

$F$ is a $N \times K$ matrix with each element as a real value. $F_{ik}$ represents the `belongingness' of node $i$ to class $k$. It is referred to as `classification function' or `feature vector'. The aim of the semi-supervised learning problem is to find $F_{.k}$ such that it is close to the labelling function and it varies smoothly over the graph. The class of node $i$ is calculated from $F$ using the following relation. Node $i$ belongs to class $k$ if $$F_{ik} > F_{ij} \quad \forall \; j \neq k.$$
The optimization problem associated with the above stated requirements can be written as: Minimize
\begin{align*}
Q(F) = 2 & \sum_{k=1}^K F_{.k}^TD^{\sigma - 1}LD^{\sigma}F_{.k}  + \mu \sum_{k=1}^{K}(F_{.k} - Y_{.k})^TD^{2\sigma - 1}(F_{.k} - Y_{.k}),
\end{align*}

\noindent where $ L= D - A$ is the Standard Laplacian of the similarity graph. It can be verified that the above problem is a convex optimization problem. The first order optimality condition gives the following solution to this problem: $$F_{.k} = (1-\alpha) (I - \alpha B)^{-1}Y_{.k},$$ where $B = D^{-\sigma}AD^{\sigma-1}$ and $\alpha = \frac{2}{2+\mu}$ for $\mu > 0$ \cite{semisup}. Rearranging the terms of the closed form solution gives the power iteration based algorithm:\\

\noindent\underline{\textbf{Algorithm 1}}:
$$ F^{t+1} = \alpha BF^{t} + (1 - \alpha) Y, $$

\noindent where $F^{t}$ is the estimate of $F$ after $t \geq 0 $ iterations.

In each iteration, the feature values for all the nodes are updated for all the classes. This algorithm is henceforth referred to as the \textit{power iteration algorithm}. Note that the (sparse) matrix-vector multiplication $BF^{t}$ is
the bottleneck of Algorithm~1. There is a number of very efficient parallel implementations of the matrix-vector
multiplications. One of the state of the art implementations is provided by NVIDIA CUDA platform \cite{BG09,Netal08,CUDA}.

Let $H$ be a diagonal matrix with elements as row sums of $B$, i.e., $$ H_{ii} = \sum_{j}B_{ij}. $$
$P$ is defined as $P = H^{-1}B$ and is used as the transition probability matrix on the graph with weighted edges. Since $P$ might be reducible, its irreducible counterpart as in the PageRank algorithm is: $$ Q = (1 - \epsilon)P + \frac{\epsilon}{N}E,$$ where $E$ is an $N \times N$ matrix with all $1$'s.


A stochastic approximation of the closed form solution gives the following sampling based algorithm. Let $X_t, t \geq 0$, be a Markov chain with transition matrix $Q$. Then for $t \geq 0$, do:\\

\noindent \underline{\textbf{Algorithm 2}}:
\begin{eqnarray*}
F_{ij}^{t+1} &=& F_{ij}^{t} + \eta_{t}I\{X_t = i\}\frac{p(i, X_{t+1})}{q(i,X_{t+1})} \left(H_{ii}F_{X_{t+1}j}^{t} - F_{ij}^{t} + \alpha Y_{ij}\right),
\end{eqnarray*}

\noindent where $\{\eta_{t}\}_{t \geq 0}$ is a positive step-size sequence satisfying $\sum_{t \geq 0} \eta_{t} = \infty$ and $\sum_{t \geq 0} \eta_{t}^{2} < \infty$.

Here $p(i, j), \; q(i,j)$ are the elements of $P$ and $Q$ respectively and the ratio $p(i, X_{t+1})/q(i, X_{t+1})$ is the (conditional) likelihood ratio correcting for the fact that we are simulating the chain with transitions governed by $Q$ and not $P$. This is a form of importance sampling. We run this update step $\forall j \in \{1,2,..,K\}$. In each iteration, the feature value of only one node is updated. This node is selected by the Markov sequence. Since we are sampling from the possible nodes each time before updating, this algorithm is henceforth referred to as the \textit{sampling algorithm}. It must be noted that a round robin policy or any other policy that samples the nodes sufficiently often provides a viable alternative. We return to this later.

From the very formulation of Algorithm~2, it is clear that this approach has immediate distributed asynchronous implementation.

\section{Convergence Analysis}
\label{sec3}
We now present proofs of convergence for the two algorithms. 
Consider the problem of finding a unique solution $x^*$ of the system $$ x = G(x) = \widetilde Bx + \widetilde Y,$$ where $c \in \mathcal{R}^{d}$ and $\widetilde B = [[b(i,j)]] \in \mathcal{R}^{d \times d}$ is irreducible non-negative with Perron-Frobenius eigenvalue $\lambda \in (0,1)$ and the corresponding (suitably normalized) positive eigenvector $w = [w_1,...w_d]^T$. Define the weighted norm $$ \|x\|_{w} = \max_{i} \left|\frac{x_j}{w_i}\right|.$$

\noindent \textbf{Lemma}
The map $G$ is a contraction w.r.t. the above norm, specifically,
$$
\|G(x) - G(y)\|_{w} \leq \lambda \|x-y\|_{w}.
$$
\\
\noindent \textbf{Proof}
We have
\begin{equation}\nonumber
\begin{aligned}
	 \|G(x) - G(y)\|_{w} & = \max_{i} \left|\frac{\sum_{j}b(i,j)(x_j - y_j)}{w_i}\right| \\
	& = \max_{i} \left|\frac{\sum_{j}b(i,j)\frac{(x_j - y_j)}{w_j}w_j}{w_i}\right| \\
	& \leq \left(\max_{j} \left|\frac{x_j - y_j}{w_j}\right|\right) \left( \max_{i} \left|\frac{\sum_{j}b(i,j)w_j}{w_i}\right|\right)\\
	& = \lambda \|x-y\|_{w}.
\end{aligned}
\end{equation}
This proves the claim. \hfill $\Box$

Identifying $\tilde{B}, \tilde{Y}$ with $\alpha B, (1 - \alpha)Y$, our Algorithm 1 reduces to the above column-wise. Algorithm 1 is then simply a repeated application of a contraction map with contraction coefficient $\lambda$ and by the celebrated contration mapping theorem, converges to its unique fixed point (i.e., our desired solution) at an exponential rate with exponent $\lambda$. Since matrix $B$ is similar to the stochastic matrix $D^{-1}A$ with one as its largest
in modulus eigenvalue, we have that $\lambda=\alpha$.

We consider Algorithm 2 next. Write $\widetilde B = \widetilde H \widetilde P$ where $\widetilde H = diag(l(1), ...l(d))^T, l(i) = \sum_{j}b(i,j)$, is a diagonal matrix and $ \widetilde P = [[\tilde{p}(i,j)]]$ is a stochastic matrix given by $\tilde{p}(i,j) = \frac{b(i,j)}{l(i)}$. Consider the scheme similar to algorithm 2:
\begin{equation}
x_{i}^{t+1} = x_{i}^{t} + \eta_{t}I\{X_t = i\}\frac{\tilde{p}(i, X_{t+1})}{\tilde{q}(i,X_{t+1})}(l(i)x_{X_{t+1}}^{t} - x_{i}^{t}), \label{alg3}
 \end{equation}
where ${X_t}$ is a Markov Chain with transition probability matrix $$\widetilde Q = [[\tilde{q}(i,j)]], \; \tilde{q}(i,j) = \frac{\epsilon}{d} + (1-\epsilon) \tilde{p}(i,j)$$.

\noindent We have the following result.

\bigskip

\noindent \textbf{Theorem}
Almost surely, $x^t \rightarrow x^*$ as $t \to \infty$.\\

 \noindent \textbf{Proof (sketch)}
Note that
 \begin{eqnarray*}
E\left[ \frac{\tilde{p}(i, X_{t+1})}{\tilde{q}(i,X_{t+1})}l(i)x_{X_{t+1}}^{t} | X_t = i\right]
&=& \sum_j \tilde{q}(i, j)\frac{\tilde{p}(i, j)}{\tilde{q}(i, j)}l(i)x_j^{t} \\
&=& \sum_jb(i, j)x^t_j.
\end{eqnarray*}
Then we can rewrite (\ref{alg3}) as
\begin{eqnarray}
x_{i}^{t+1} &=& x_{i}^{t} + \eta_{t}I\{X_t = i\}\Big(\sum_jb(i, j)x^t_j \nonumber
 - \ x_{i}^{t} +  M^{t+1}_i\Big), \label{alg4}\\
\mbox{where} \ M^{t+1}_i &:=& \Big[\frac{\tilde{p}(i, X_{t+1})}{\tilde{q}(i,X_{t+1})}(l(i)x_{X_{t+1}}^{t}) - \sum_jb(i, j)x^t_j\Big] \nonumber
 \end{eqnarray}
defines a martingale difference sequence (i.e., a sequence of integrable random variables uncorrelated with the past).  Iteration (\ref{alg4}) is a stochastic approximation algorithm that can be analyzed using the `ODE' (for \textit{Ordinary Differential Equation}') approach which treats the discrete time algorithm as a noisy discretization (or `Euler scheme') of a limiting ODE with slowly decreasing step-size, decremented at just the right rate so that the errors due to discretization and noise are asymptotically negligible and the correct asymptotic behavior of the ODE is replicated \cite{book}. See p.\ 74-75, Chapter 6 of \textit{ibid.}  for a precise statement. In particular, if the ODE has a globally asymptotically stable equilibrium, the algorithm converges to it with probability one under mild conditions stated in \textit{ibid.} which are easily verified in the present case. Let $ \pi = [ \pi(1),..,,\pi(d)]^T$ denote the stationary distribution under $\widetilde Q$. Let $\Gamma = diag(\pi(1),...,\pi(d))$. The limiting ODE in this case can be derived as in Chapter 6, \cite{book} and is
\begin{equation}\nonumber
\begin{aligned}
	\dot{x}(t) & = \Gamma(G(x(t)) - x(t))\\
	& = \widetilde{G}(x(t)) - x(t),
\end{aligned}
\end{equation}
where $$ \widetilde{G}(x) = (I - \Gamma)x + \Gamma G(x).$$
Now,
\begin{equation}\nonumber
\begin{aligned}
\|\widetilde{G}(x) - \widetilde{G}(y)\|_{w}
&\leq \max_i \left[(1-\pi(i))\left|\frac{x_i - y_i}{w_i}\right| + \pi(i)\left|\frac{\sum_{j}b(i,j)(x_j - y_j)}{w_i}\right|\right] \\
& \leq \max_i \left[(1-\pi(i))\left|\frac{x_i - y_i}{w_i}\right| + \pi(i)\lambda\|x - y\|_{w} \right]\\
& \leq \tilde{\lambda}\|x-y\|_{w},
\end{aligned}
\end{equation}
where, $$ \tilde{\lambda} = \max_{i}(1 - \pi(i) + \lambda\pi(i)) \in (0,1). $$
Thus, $\widetilde{G}$ is also a contraction w.r.t $\|.\|_{w}$. By results of p.\ 126-127 of \cite{book}, the ODE converges to the unique fixed point of $\widetilde{G}$, equivalently of $G$, i.e. the desired solution $x^*$. Furthermore, if we consider the scaled limit $\tilde{G}_{\infty}(x) := \lim_{a\uparrow\infty}\frac{\tilde{G}(ax)}{a}$ (which corresponds to replacing  $\tilde{Y}$ by the zero vector in the above), then the ODE $\dot{y}(t) = \tilde{G}_{\infty}(y(t)) - y(t)$ is seen to have the origin as its unique globally asymptotically stable equilibrium. By Theorem 9, p.\ 75, \cite{book}, $\sup_t\|x^t\| < \infty$ with probability one. In other words, the iteration is stable. Hence by the theory developed in Chapter 2 (particularly Theorems 7 and Corollary 8, p.\ 74) of \textit{ibid.}, $\{x^t\}$ converges to $x^*$ with probability one. \hfill  $\Box$

In Algorithm $2$, if we replace $ \alpha B$ by $\widetilde B$, then the algorithm is column-wise exactly the same as (\ref{alg3}). Hence Algorithm 2 converges to the correct value of the feature matrix $F$. Note the key role of the factor $\alpha$ in ensuring the contraction property.

If we consider round robin sampling, the results of \cite{book}, Section 4.2 on sample complexity become applicable here. In particular, we conclude that at time $n$, the probability of remaining within a prescribed small neighborhood of $x^*$ after $n + \tau$ iterates (where $\tau$ can be prescribed in terms of problem parameters) remains greater than $1 - O\left(e^{\frac{C}{\sum_{t \geq n}\eta_t^2}}\right)$. If for example, $\eta_t = \Theta\left(\frac{1}{t}\right)$, then $\sum_{t \geq n}\eta_t^2 = \Theta\left(\frac{1}{t}\right)$ and this implies exponential decay of probability of ever escaping from this neighborhood after $n + \tau$ iterations. We may expect a similar behavior for Markov sampling, but analogous estimates for asynchronous stochastic approximation with `Markov' noise (of which Markov sampling is a special case) appear to be currently unavailable.

\section{Experiments and Results}
\label{sec4}
To evaluate the classification accuracy of the two algorithms, we have used the following networks (real and synthetic): the graph based on the French classic Les Miserables written by Victor Hugo \cite{lesmis}, the graph of university webpages dataset as found on WebKB \cite{webKB}, Gaussian mixture graphs, and finally dynamic stochastic block model graph based on M/M/K/K queue. The number of updates in the power method in one iteration is $N$ while the classification function of only one node is updated in each step of the sampling algorithm. Hence we shall refer to $N$ steps of the sampling algorithm as one iteration, to keep the comparison fair. The number of misclassified nodes is a measure of error and is used as the metric for performance testing. It must be noted that the rate of decrease of the step size also influences the rate of convergence of the error.

\subsection{Les Miserables graph}

Characters of the novel Les Miserables form the nodes of the graph \cite{lesmis}. The nodes that appear on the same page in the novel are connected by a weighted directed edge. We will consider the undirected version of this graph with the weight of all the nodes set to $1$. This graph has $77$ nodes and $6$ clusters, namely Myriel ($10$), Valjean ($17$), Fantine ($10$), Cosette ($10$), Thenardier ($12$), Gavroche ($18$) where the class name is the name of a character and in the bracket is indicated the number of nodes in that cluster. Class information of the nodes named above is available to the algorithms. We use the decreasing step size $\frac{1}{2 + \left \lfloor{t/100}\right \rfloor}$.

\begin{figure}[h]
\centering
\begin{subfigure}[a]{0.49\linewidth}
\centering
\includegraphics[width=\linewidth]{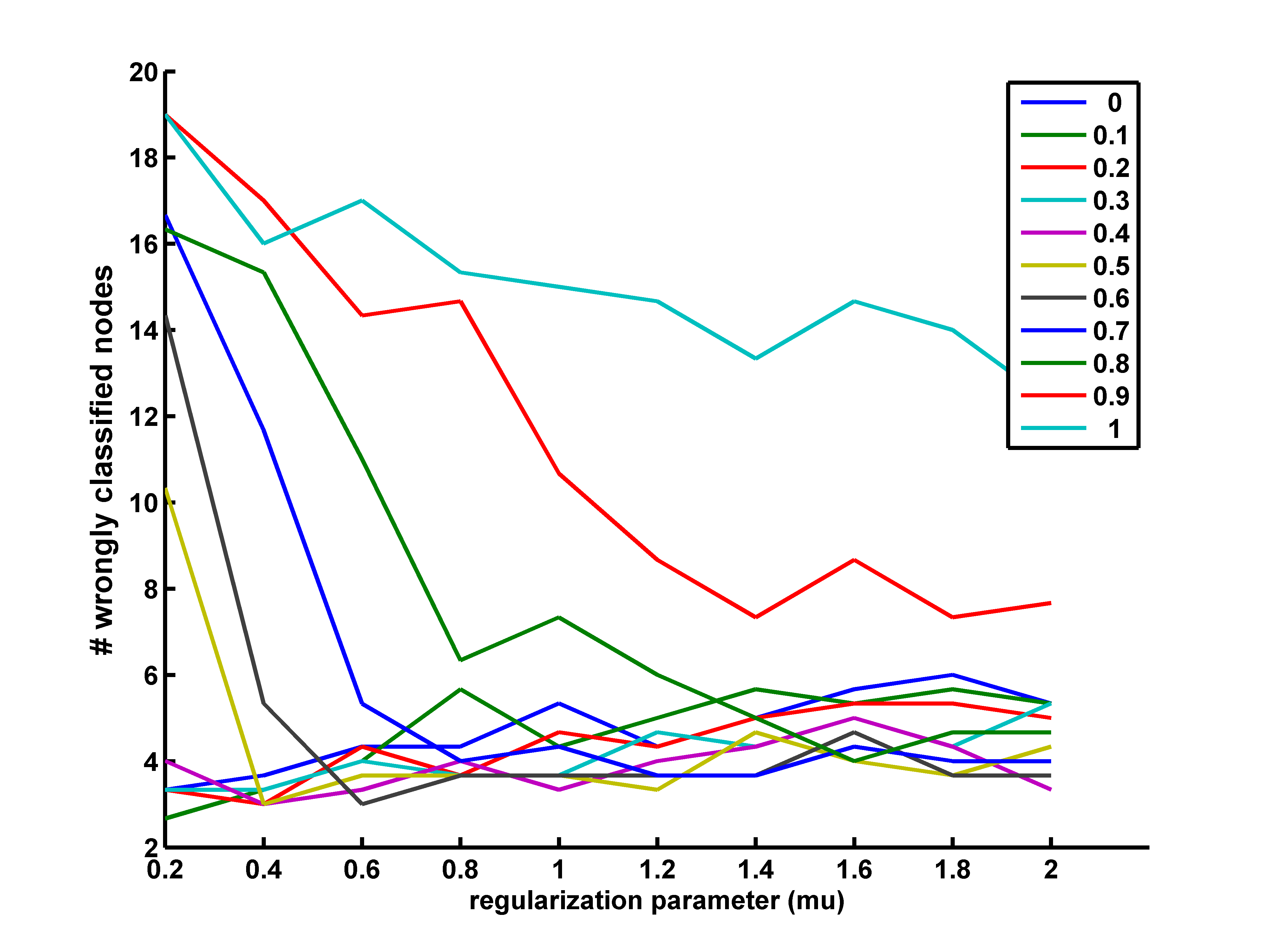}
\captionof{figure}{The error for different values of $\sigma$ and $\mu$.}
\end{subfigure}
\hfill
\begin{subfigure}[a]{0.49\linewidth}
\centering
\includegraphics[width=\linewidth]{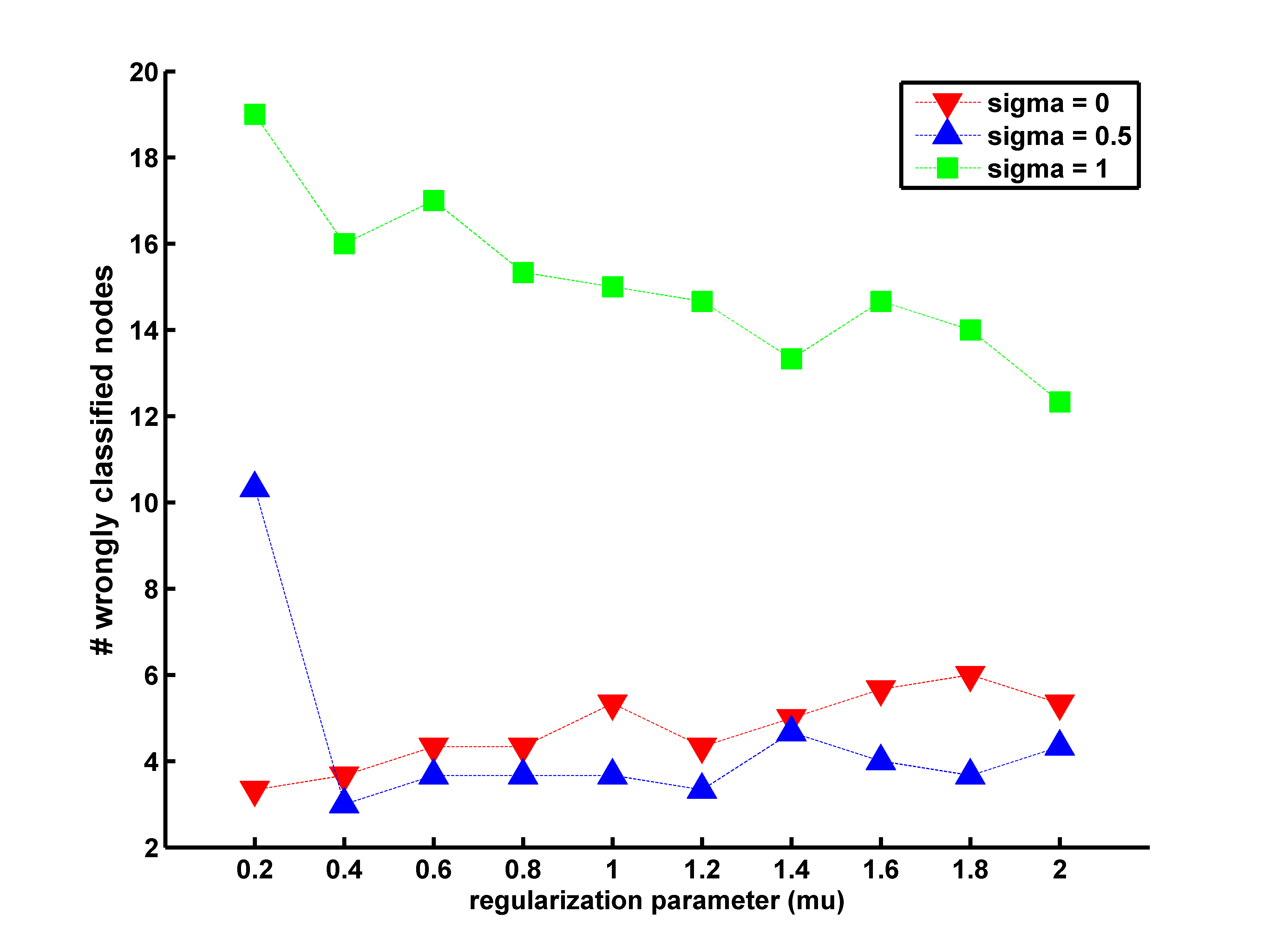}
\captionof{figure}{The error shown specifically for $\sigma = 0, 0.5, 1$.}
\end{subfigure}
\caption{Error for different values of $\mu$ and $\sigma$.}
\label{1}
\end{figure}

Algorithms $1$ and $2$ provide $\mu$ and $\sigma$ as two parameters that can be varied. For different values of these parameters, Algorithm $1$ was run for $500$ iterations and then the class of each node was found. This was done $3$ times for each value of $\mu$ and $\sigma$ and then the average error across the $3$ classifications was calculated. The average error is shown in \figurename{1} (a). 
\figurename{1}(b) shows the same information for the values of $\sigma$ corresponding to the Standard Laplacian ($\sigma = 1$), Normalised Laplacian ($\sigma = 0.5$) and PageRank ($\sigma = 0$).

For further experiments, we will be considering the value $\sigma = 0.5$ and $\mu = 1$.
According to \figurename{1} this is a good choice of the parameters. This choice of the parameters can also
be backed up by the results from \cite{AGS13} and \cite{Zetal04}.
Note that $\sigma = 0.5$ corresponds to the Normalised Laplacian algorithm. The difference in performance for $\sigma$'s can be understood by taking the analogy of a random walk on the graph starting from the labelled node of the cluster \cite{AGS13}. In general, if there is a node that is connected to the labelled node of two clusters, one of which is denser with the labelled node having degree higher than the other labelled node and the other cluster is less dense or smaller, then the PageRank algorithm tends to classify it into the smaller cluster while the Standard Laplacian classifies it into the denser cluster. 

\begin{figure}[h]
\centering
\begin{subfigure}[a]{0.49\linewidth}
\centering
\includegraphics[width=\linewidth]{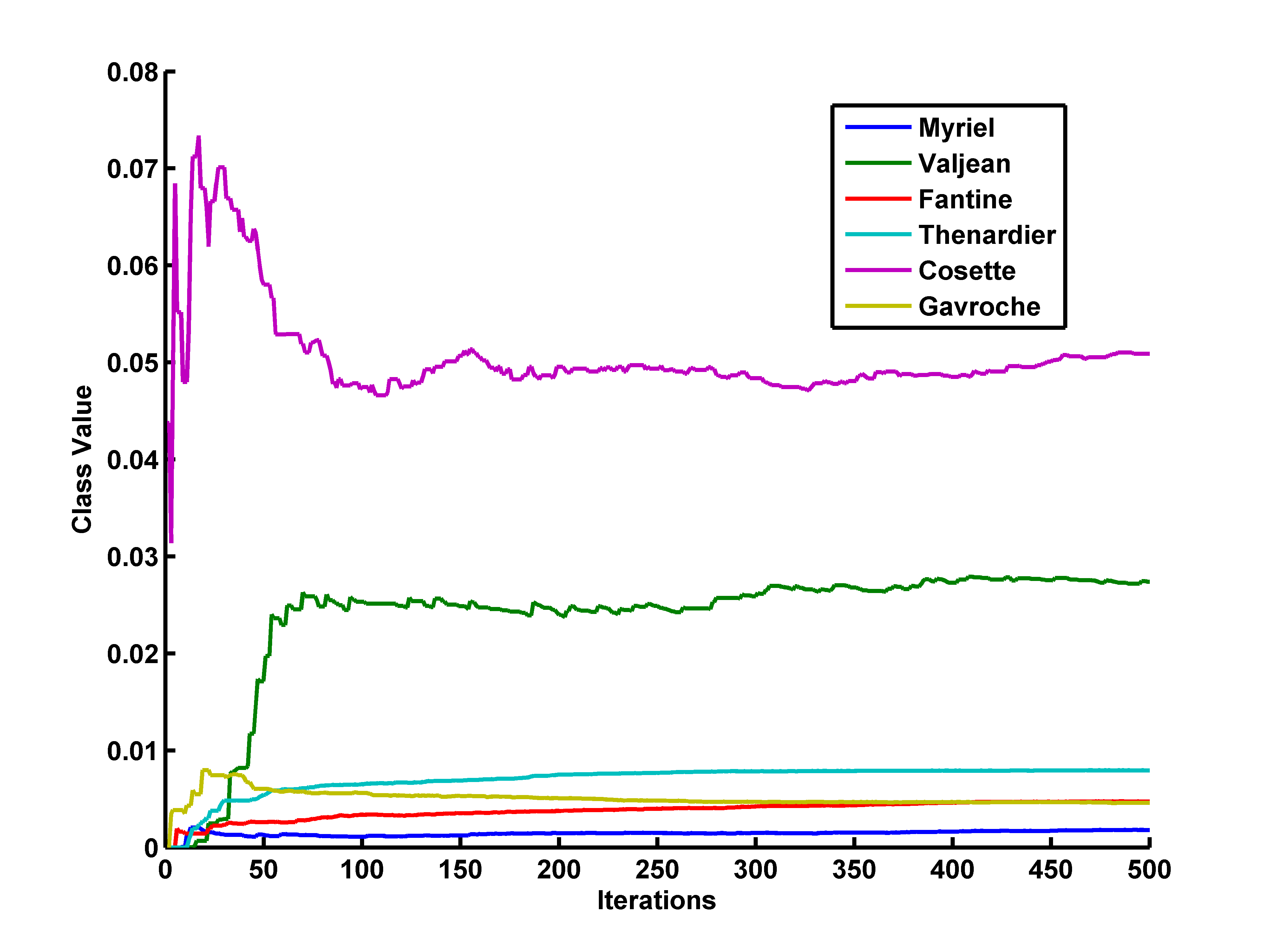}
\captionof{figure}{Sampling Algorithm}
\end{subfigure}
\hfill
\begin{subfigure}[a]{0.49\linewidth}
\centering
\includegraphics[width=\linewidth]{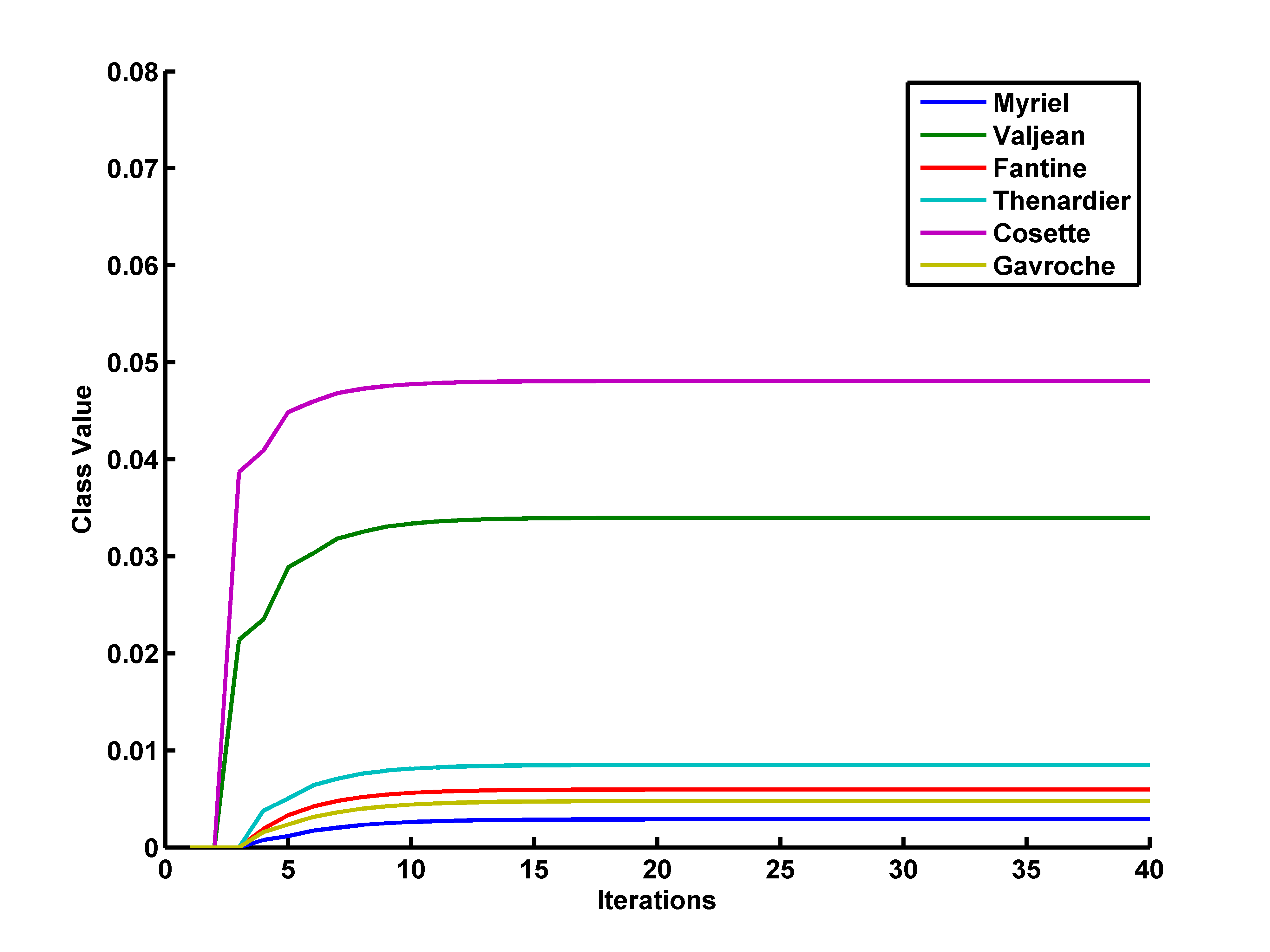}
\captionof{figure}{Power Iteration Algorithm}
\end{subfigure}
\caption{Plots for the node Woman2. In both cases, it is classified into the class \textit{Cosette}.}
\label{2}
\end{figure}

\begin{figure}[h]
\centering
\begin{subfigure}[a]{0.49\linewidth}
\centering
\includegraphics[width=\linewidth]{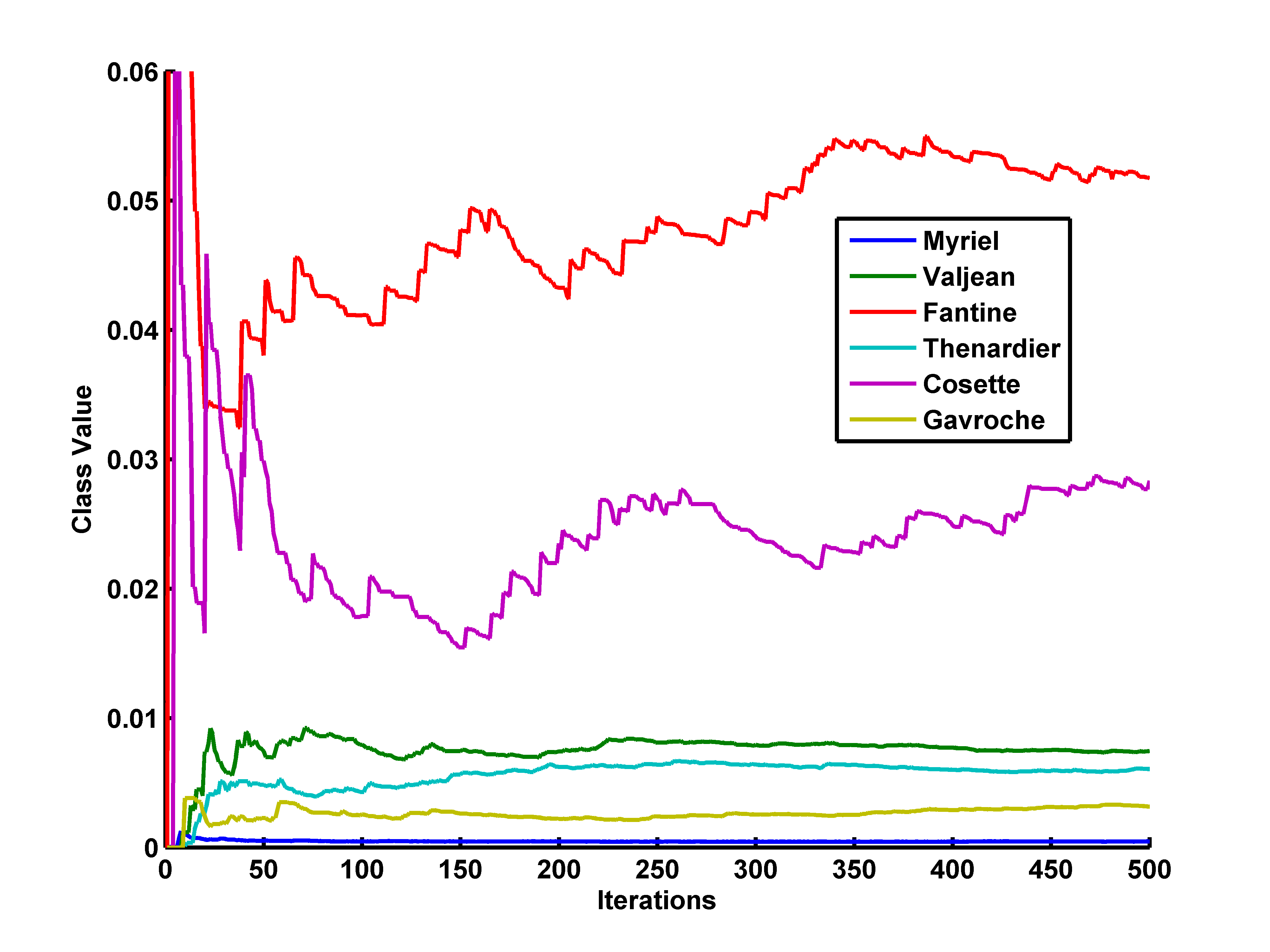}
\captionof{figure}{Sampling Algorithm}
\end{subfigure}
\hfill
\begin{subfigure}[a]{0.49\linewidth}
\centering
\includegraphics[width=\linewidth]{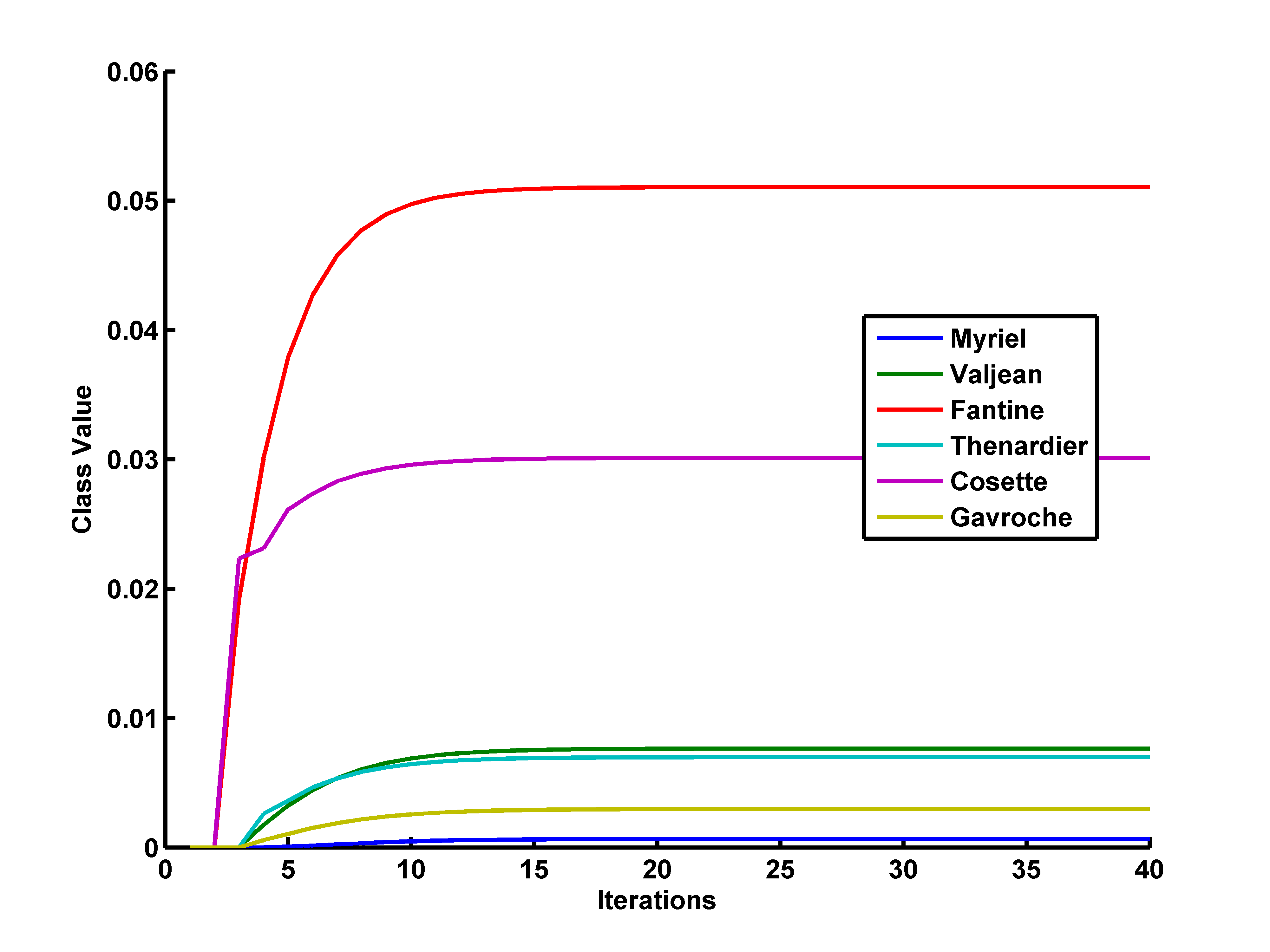}
\captionof{figure}{Power Iteration Algorithm}
\end{subfigure}
\caption{Plots for the node Tholomyes. In both cases, it is classified into the class \textit{Fantine}.}
\label{3}
\end{figure}

\begin{figure}[h]
\centering
\begin{subfigure}[a]{0.49\linewidth}
\centering
\includegraphics[width=\linewidth]{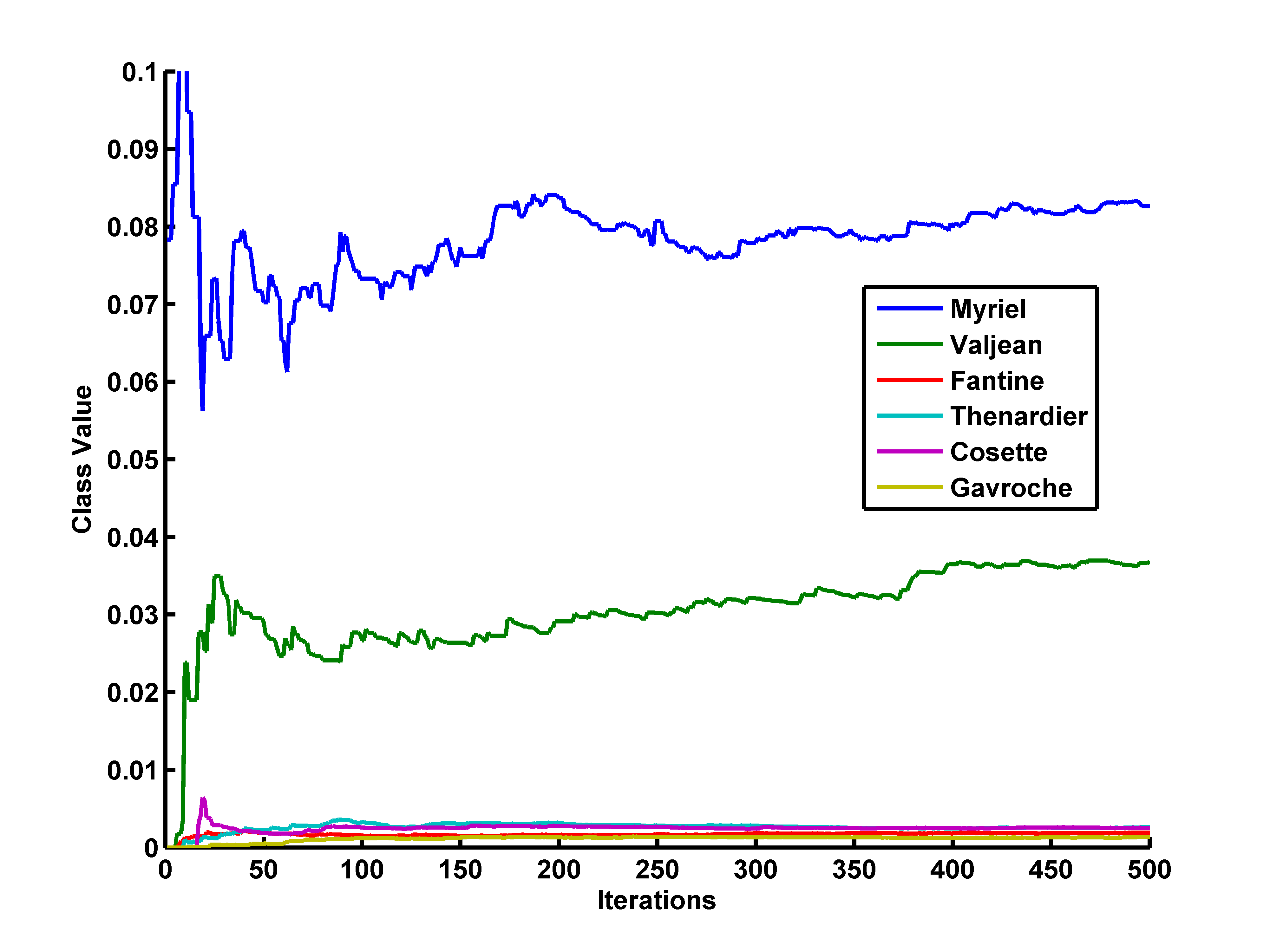}
\captionof{figure}{Sampling Algorithm}
\end{subfigure}
\hfill
\begin{subfigure}[a]{0.49\linewidth}
\centering
\includegraphics[width=\linewidth]{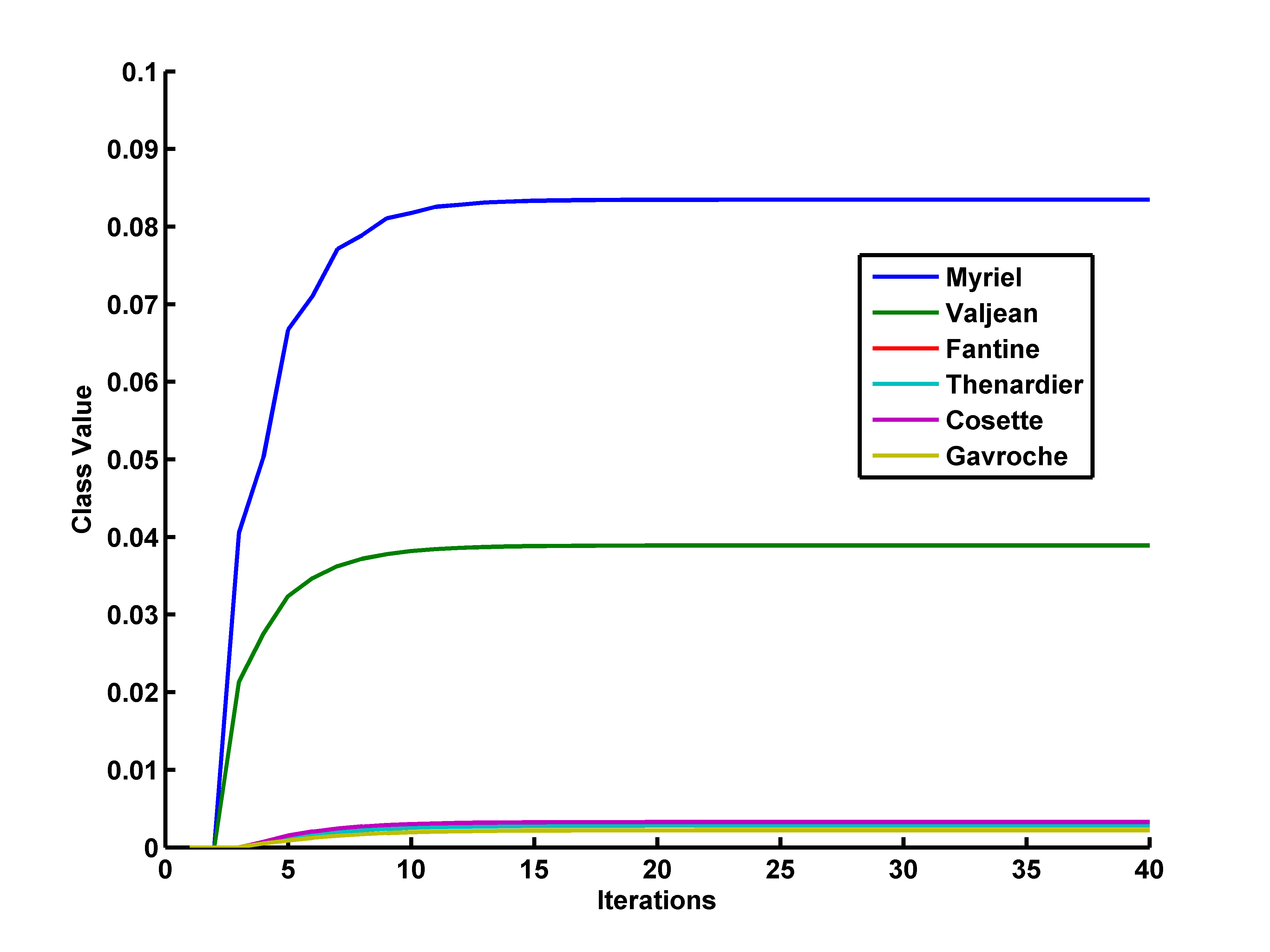}
\captionof{figure}{Power Iteration Algorithm}
\end{subfigure}
\caption{Plots for the node MlleBaptistine. In both cases, it is classified into the class \textit{Myriel}.}
\label{4}
\end{figure}

The convergence of classification function for the two algorithms can be seen in \figurename{2}, \figurename{3}, \figurename{4}. It can also be seen that the classification function for the sampling algorithm converges to the same values as the classification function for the power iteration algorithm.

\begin{figure}[h]
\begin{center}
\includegraphics[scale=0.45]{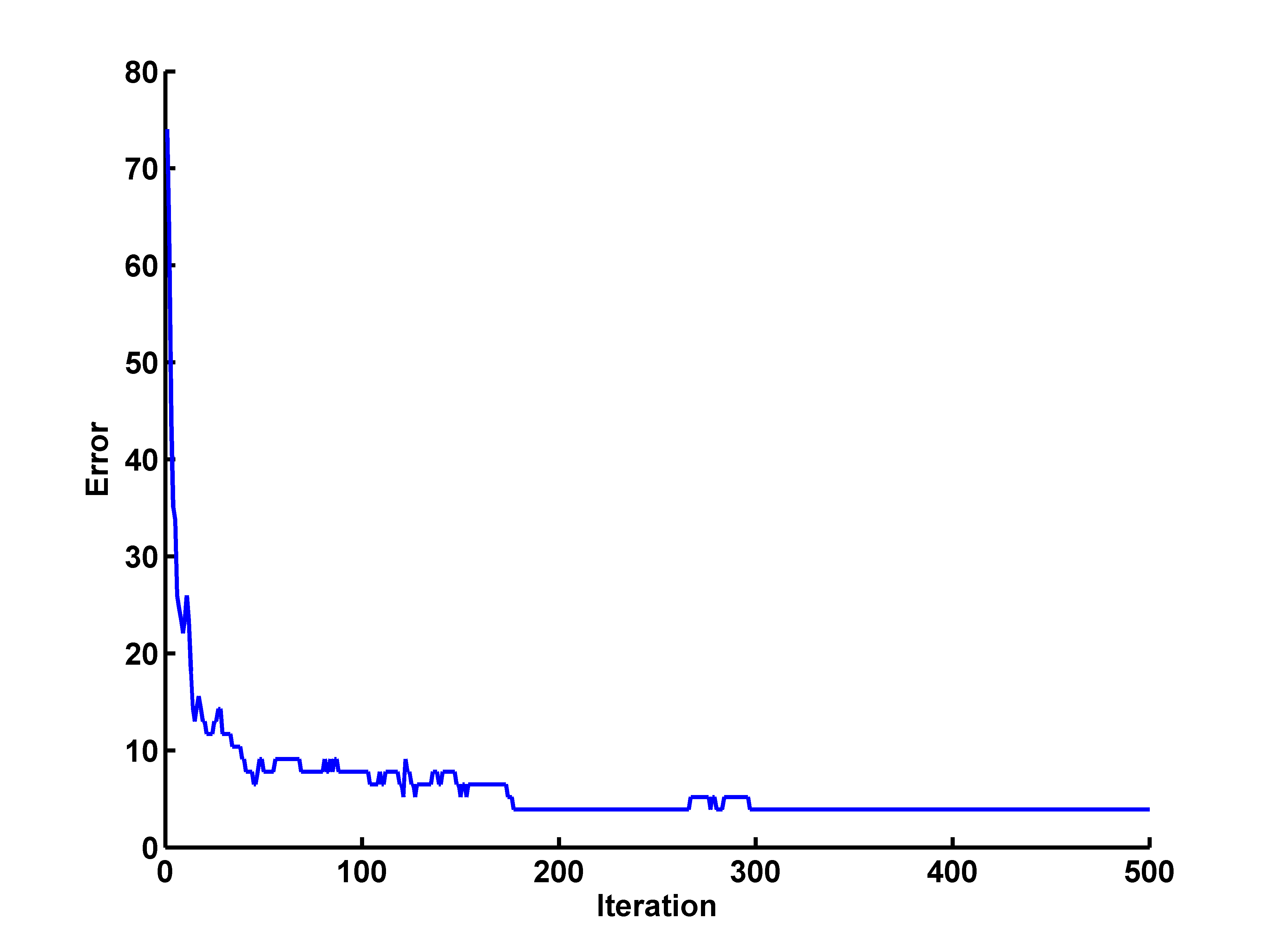}
\caption{Error evolution for Les Miserables graph with $\sigma = 0.5$ and $\mu = 1$.}
\label{5}
\end{center}
\end{figure}

Les Miserables graph has interesting nodes like Woman2, Tholomyes and Mlle Baptistine among others which are connected to multiple labelled nodes. Since the class density and the class sizes for different nodes is different, these nodes get classified into different classes for different values of $\sigma$. The node Woman2 is connected to the labelled nodes Cosette, Valjean and Javert(which belongs to class Thenardier). As seen in the \figurename{3}, the classification function values for these nodes/classes is higher compared to the other $3$ classes. Similar is the case for Tholomyes as shown in \figurename{4} and Mlle Baptistine as shown in \figurename{4}. The error as a function of iteration number is shown in \figurename{5}.

%

%

\subsection{WebKB Data}
We next see the classification of webpages of $4$ universities - Cornell, Texas, Washington and Wisconsin - corresponding to the popular WebKB dataset \cite{webKB}. We considered the graph formed by the hyperlinks connecting these pages. Webpages in the original dataset that do not have hyperlinks to webpages within the dataset are not considered. This gives the following clusters Cornell (677), Texas (591), Washington (983) and Wisconsin (616). The node with the largest degree is selected from each class (university) as the labelled node. We use the decreasing step size $\frac{1}{2 + \left \lfloor{t/1000}\right \rfloor}$. We will be using this decreasing step size for the other experiments unless stated otherwise.

\begin{figure}
\begin{center}
\includegraphics[scale=0.45]{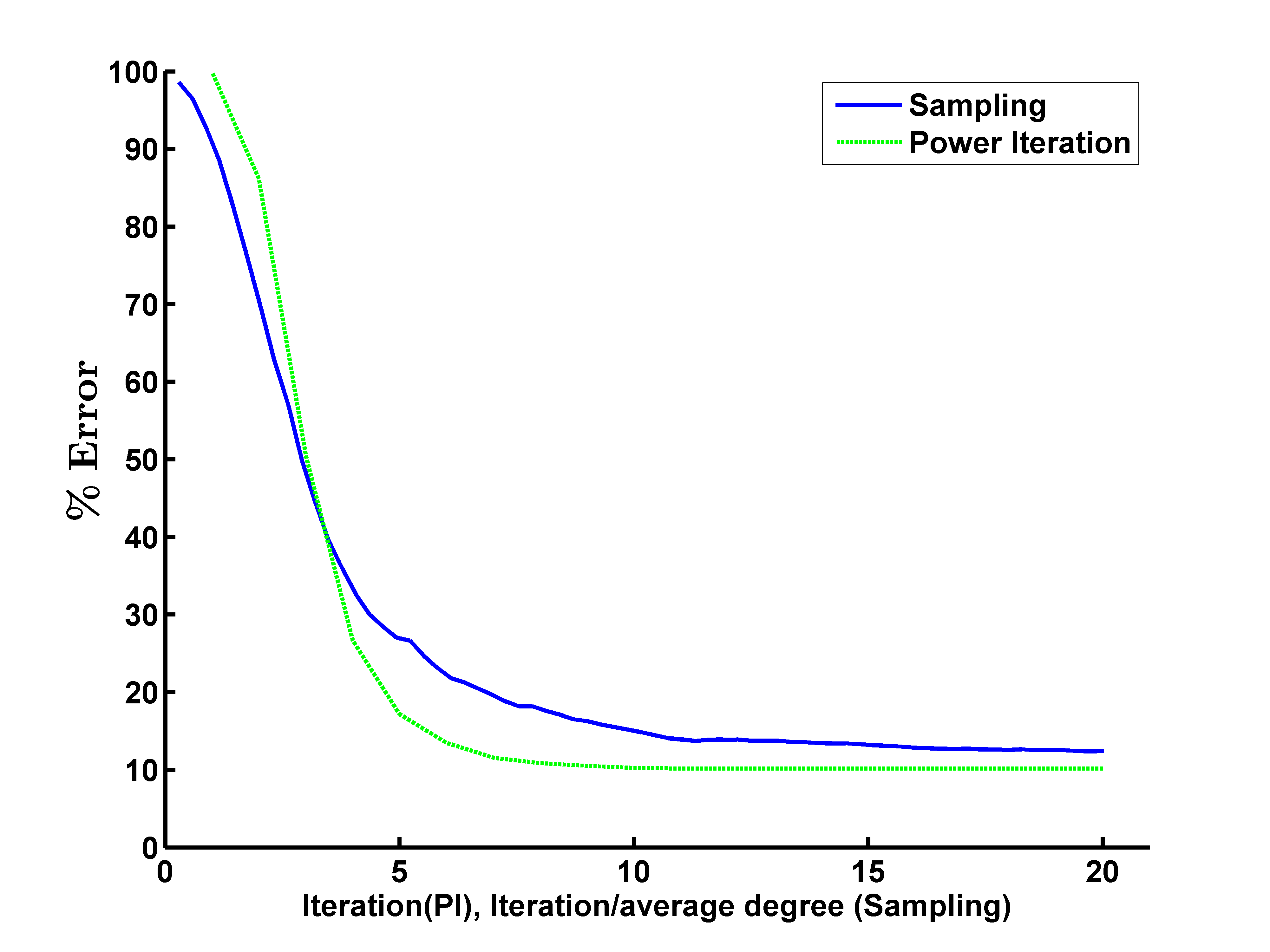}
\caption{Error evolution for the WebKB graph with 4 universities as classes and webpages as nodes.}
\label{6}
\end{center}
\end{figure}

\figurename{6} shows the $\%$error evolution. Majority of this error is because of the nodes belonging to Wisconsin getting classified as nodes belonging to Cornell. The degree of the labelled node of class Cornell is almost thrice that of class Wisconsin, while the number of nodes in these classes is approximately the same. The average degree, and hence the density, of the class corresponding to Wisconsin (3.20) is lesser as compared to the others (3.46, 3.54, 3.52). Iterations/average degree is used as the x-axis for the sampling algorithm in \figurename{6} while iterations is used for the power iteration algorithm. The rationale for dividing by the average degree is that the sampling algorithm obtains information from only one neighbour in one step while the power iteration obtains information from all the neighbours in one iteration. Division by average degree makes the two methods comparable in computational cost. As can be seen, the two methods are comparable in terms of accuracy as well as the computational cost for attaining a given accuracy.

It is interesting to observe that if one would like to obtain quickly a very rough classification, the sampling algorithm
might be preferable. This situation is typical for MCMC type methods.

\subsection{Gaussian Mixture Models}
\begin{figure}[h]
\centering
\begin{subfigure}[a]{0.49\linewidth}
\centering
\includegraphics[width=\linewidth]{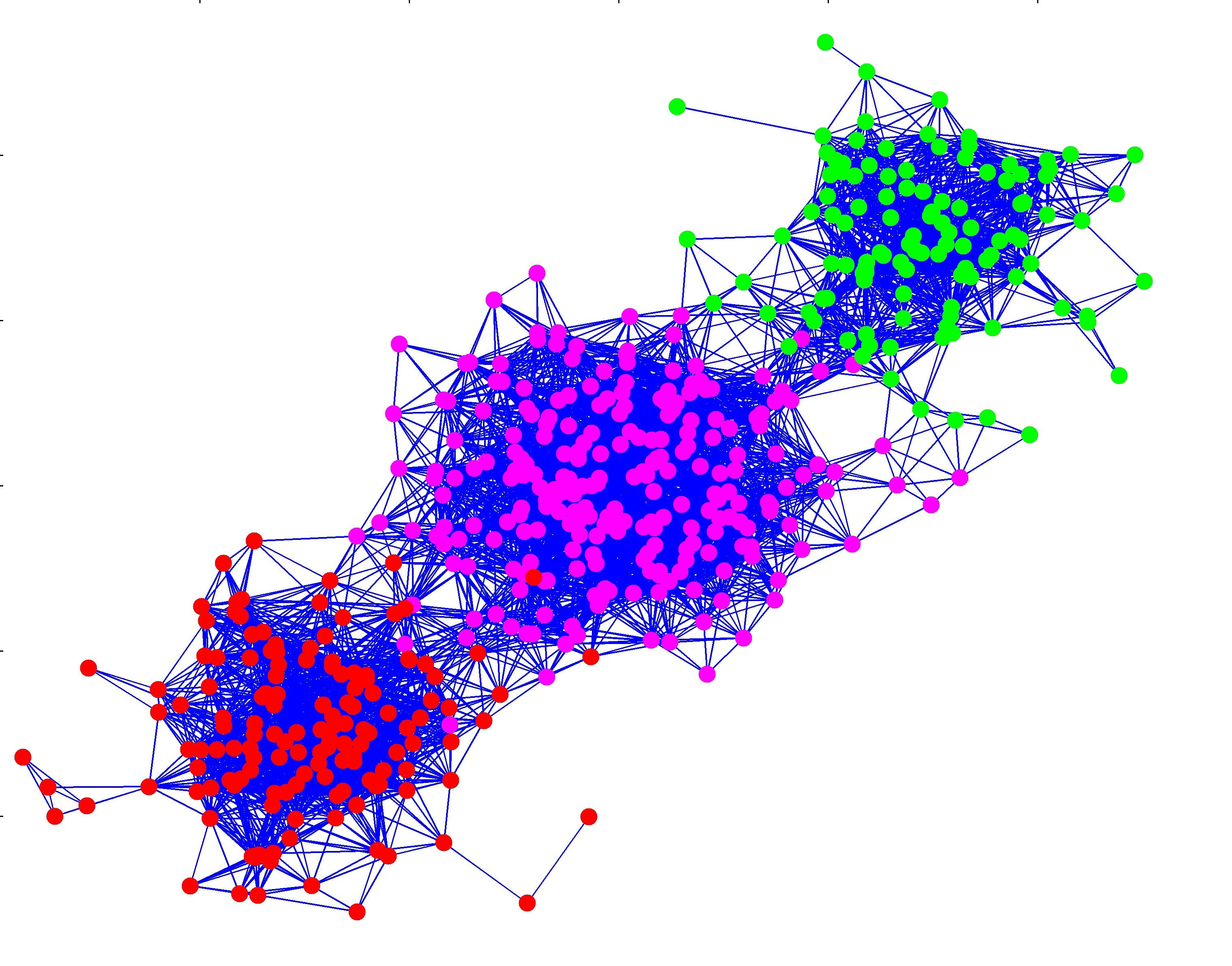}
\captionof{figure}{Graph with class of node indicated by the node's colour.}
\end{subfigure}
\hfill
\begin{subfigure}[a]{0.49\linewidth}
\centering
\includegraphics[width=\linewidth]{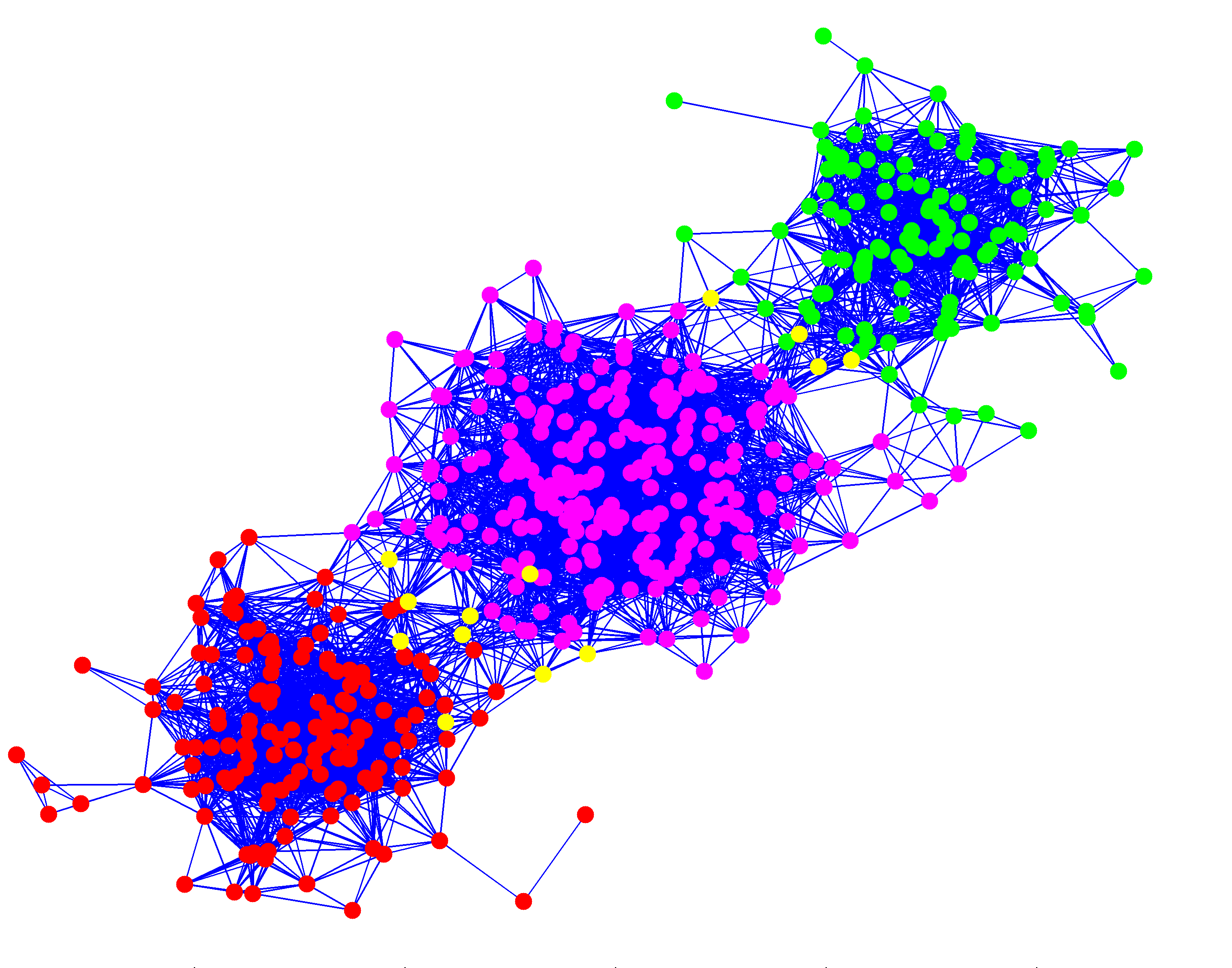}
\captionof{figure}{Graph with class of node as found by the sampling algorithm indicated by the node's colour. Misclassified nodes are shown in yellow.}
\end{subfigure}
\caption{Gaussian Mixture Graphs}
\label{7}
\end{figure}

Classification into clusters in a Gaussian Mixture graph was also tested. A Gaussian mixture graph with $3$ classes was created. A node belongs to class $1$ with probability $0.33$, to class $2$ with probability $0.33$ and to class $3$ with probability $0.34$. A node shares an edge with all the nodes within a given radius of itself. Two nodes with the highest degree from each class were chosen as the labelled nodes while the remaining nodes were unlabelled. It is known that choosing high degree nodes as labelled nodes is
beneficial \cite{AGS13}. A Gaussian mixture graph with $500$ nodes was created with the above parameters.
\figurename{7}(a) shows the graph with the nodes coloured according to their class - nodes of class $1$ are shown in pink, class $2$ in green and Class $3$ in red. \figurename{7}(b) shows the classes that the nodes were classified into using the sampling based algorithm. Nodes classified into Class $1$ are shown in pink, Class $2$ in green and Class $3$ in red. Shown in yellow are the nodes that were wrongly classified.
The nodes that are misclassified lie either on the boundary of two classes or are surrounded completely by nodes of the neighbouring class as seen in \figurename{7}(b). When this happens, the number of neighbours the node has of the other class is more than the neighbours it has of its own class, leading to misclassification.

\begin{figure}[h]
\begin{center}
\includegraphics[scale=0.45]{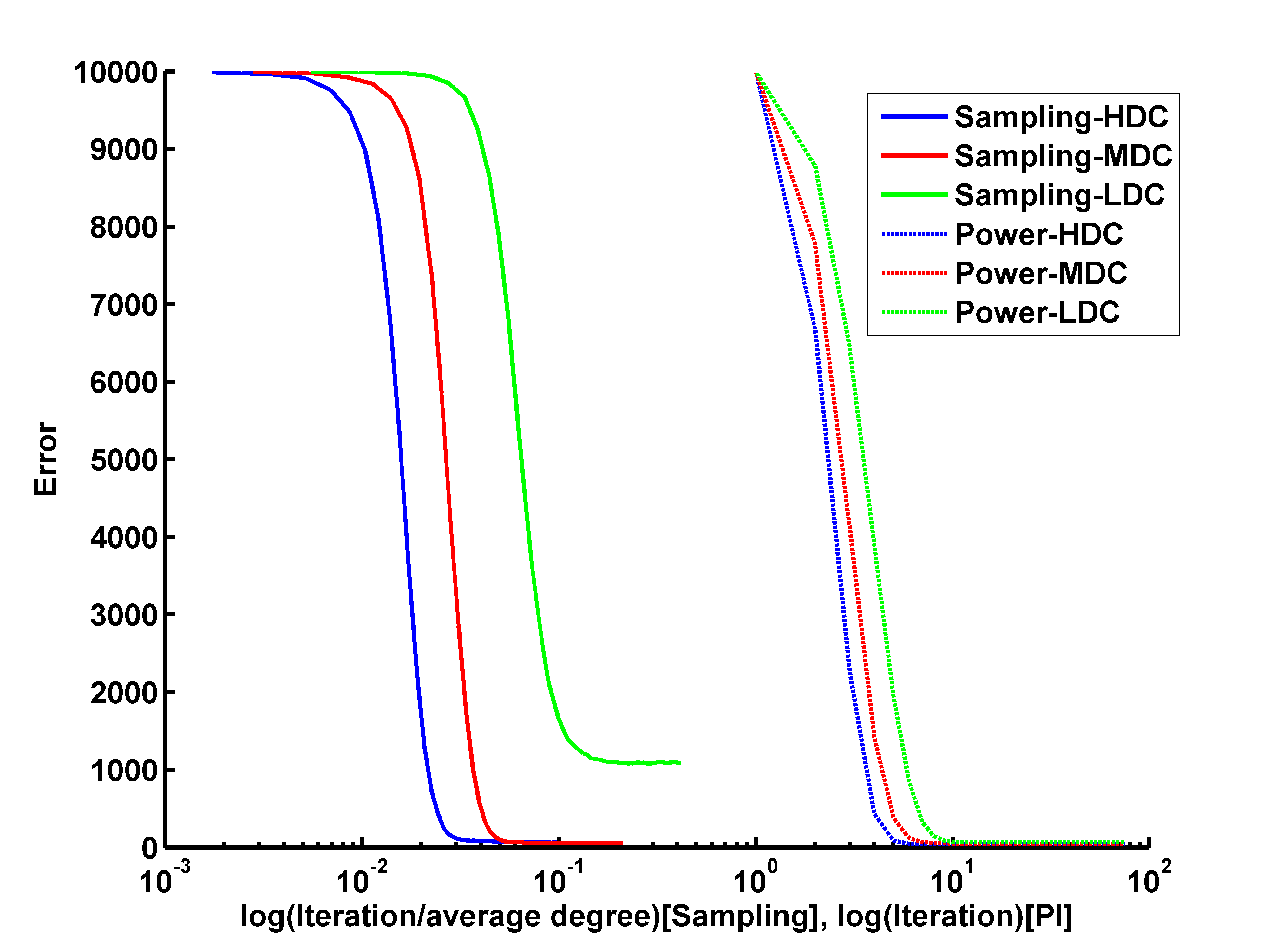}
\caption{The figure shows the error as the number of iterations increase for a graph with $10000$ nodes for both the methods. The graph with denser clusters has lesser misclassified nodes compared to the graph with less denser cluster.}
\label{8}
\end{center}
\end{figure}

The sampling algorithm scales very well which can be seen in \figurename{8}. Gaussian mixture graphs with $10000$ nodes were generated and the distributed sampling scheme was applied to them. The error decreases as the number of iterations increase, till it converges to a constant value. \figurename{8} shows the error evolution for three Gaussian graphs with different cluster densities - High Cluster Density (HCD), Medium Cluster Density (MCD), Low Cluster Density(LCD). The convergence is shown for the sampling based algorithm as well as the power iteration method. On the X-axis, the log of iterations/average degree is used for sampling method while log of iterations is used for power iteration; intention being the same as in \figurename{6}. The sampling algorithm outperforms the power iteration method in terms of computational cost required to reach a given accuracy. The number of iterations for the two methods is of the same order(5-8 for PI while 15-20 for sampling) and average node degree is between 180-570. The convergence is faster in the graph with higher cluster density. Also, the final error is smaller in graphs with higher cluster density compared to others for both methods. The misclassified nodes lie on the boundary of two classes.

\begin{figure}[h]
\begin{center}
\includegraphics[scale=0.45]{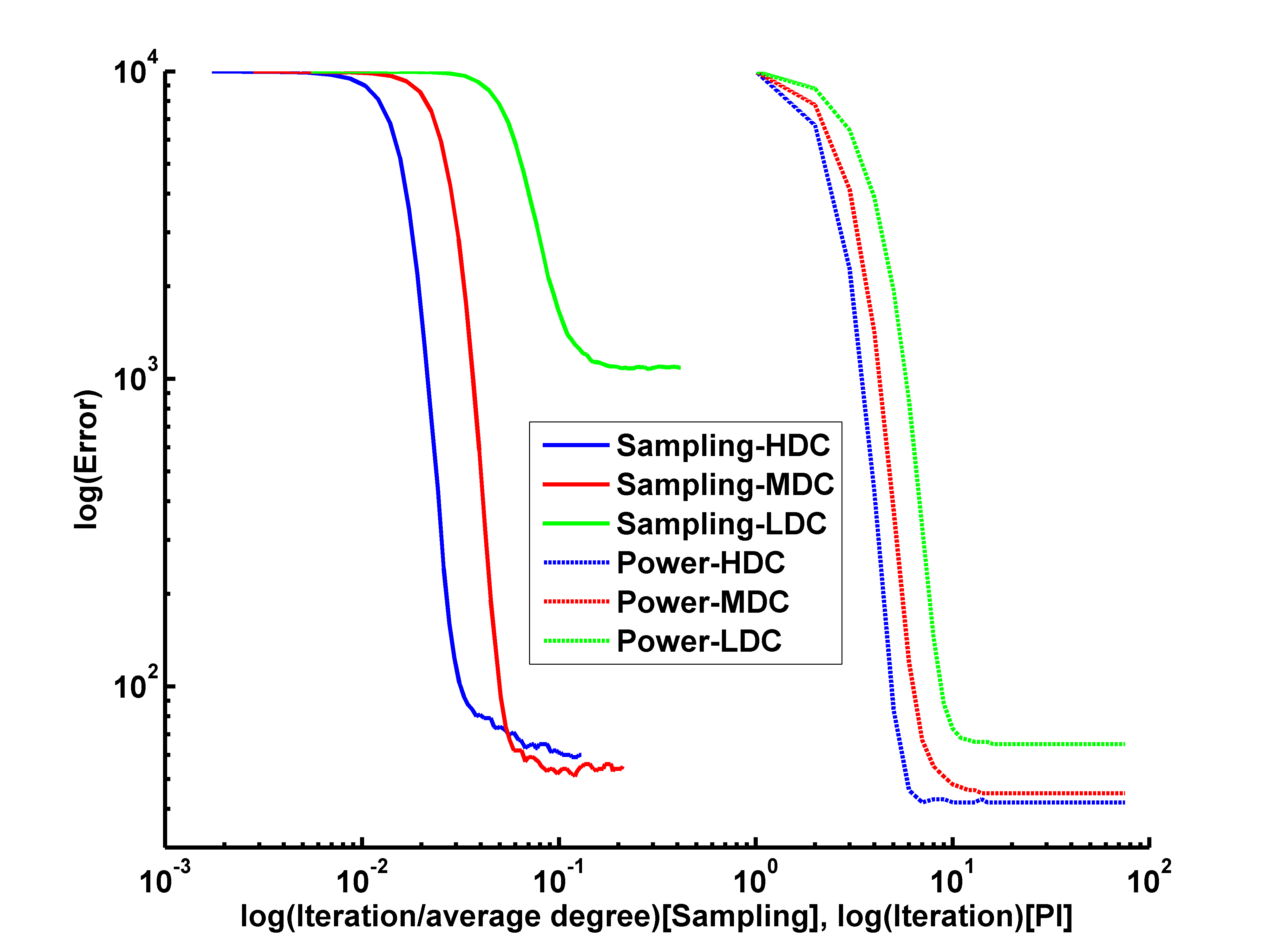}
\caption{This shows the log of error v/s the log of iterations/average degree for the HCD, MCD and LCD Gaussian mixture graphs with $10000$ nodes.}
\label{9}
\end{center}
\end{figure}


\begin{figure}[h]
\begin{center}
\includegraphics[scale=0.4]{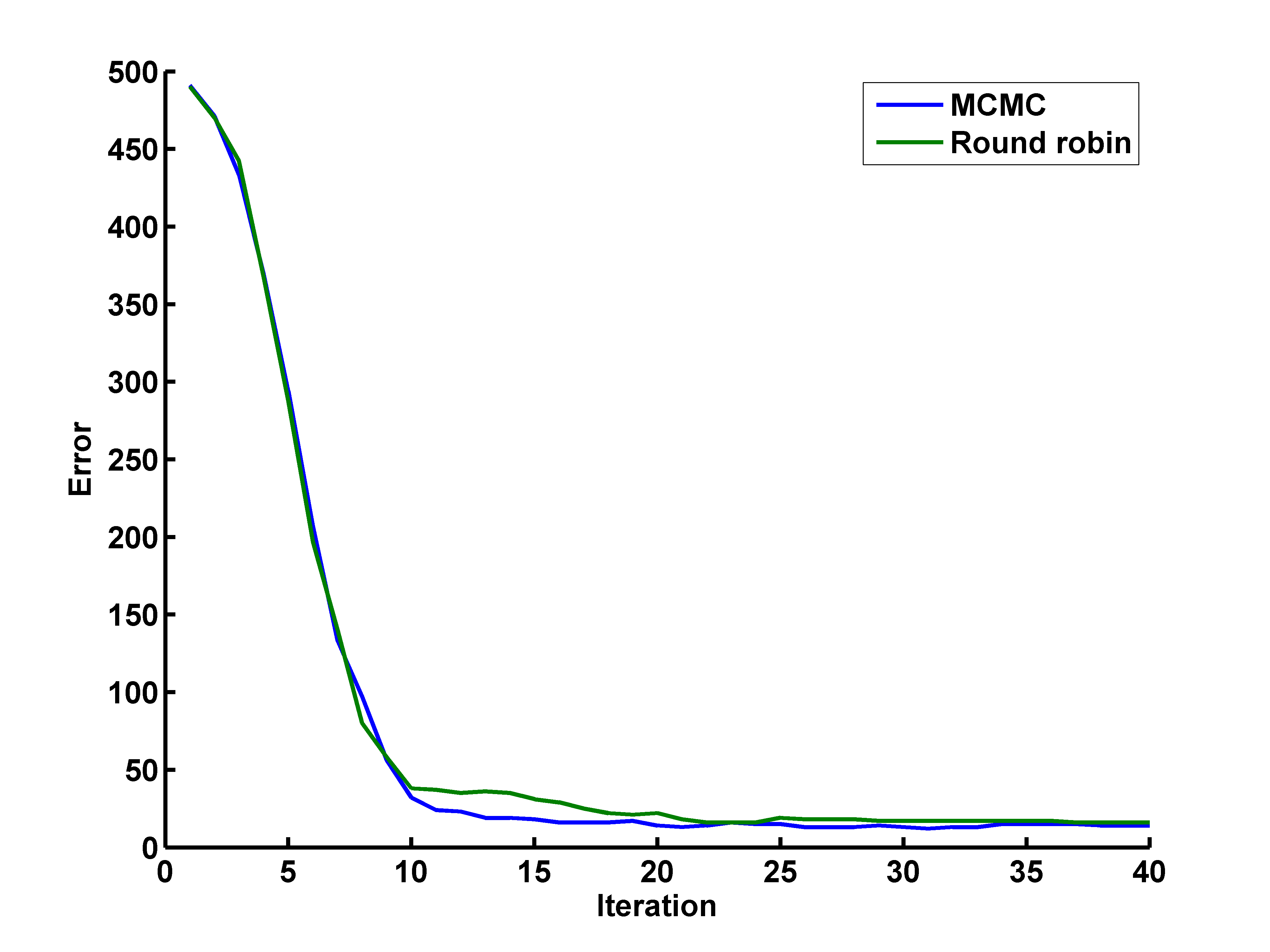}
\caption{Error evolution for Round Robin node selection in comparison with MCMC node selection.}
\label{10}
\end{center}
\end{figure}

A round robin node selection method was also tried instead of the MCMC based selection in the sampling algorithm. Node for which update was to be made was selected in a round robin fashion and then one of neighbors was sampled. It can be seen in \figurename{10} that the performance of the two node selection methods is comparable in terms of number of iterations for convergence as well as the error on convergence.

%

\subsection{Tracking of new incoming nodes}
\begin{figure}[h]
\begin{center}
\includegraphics[scale=0.45]{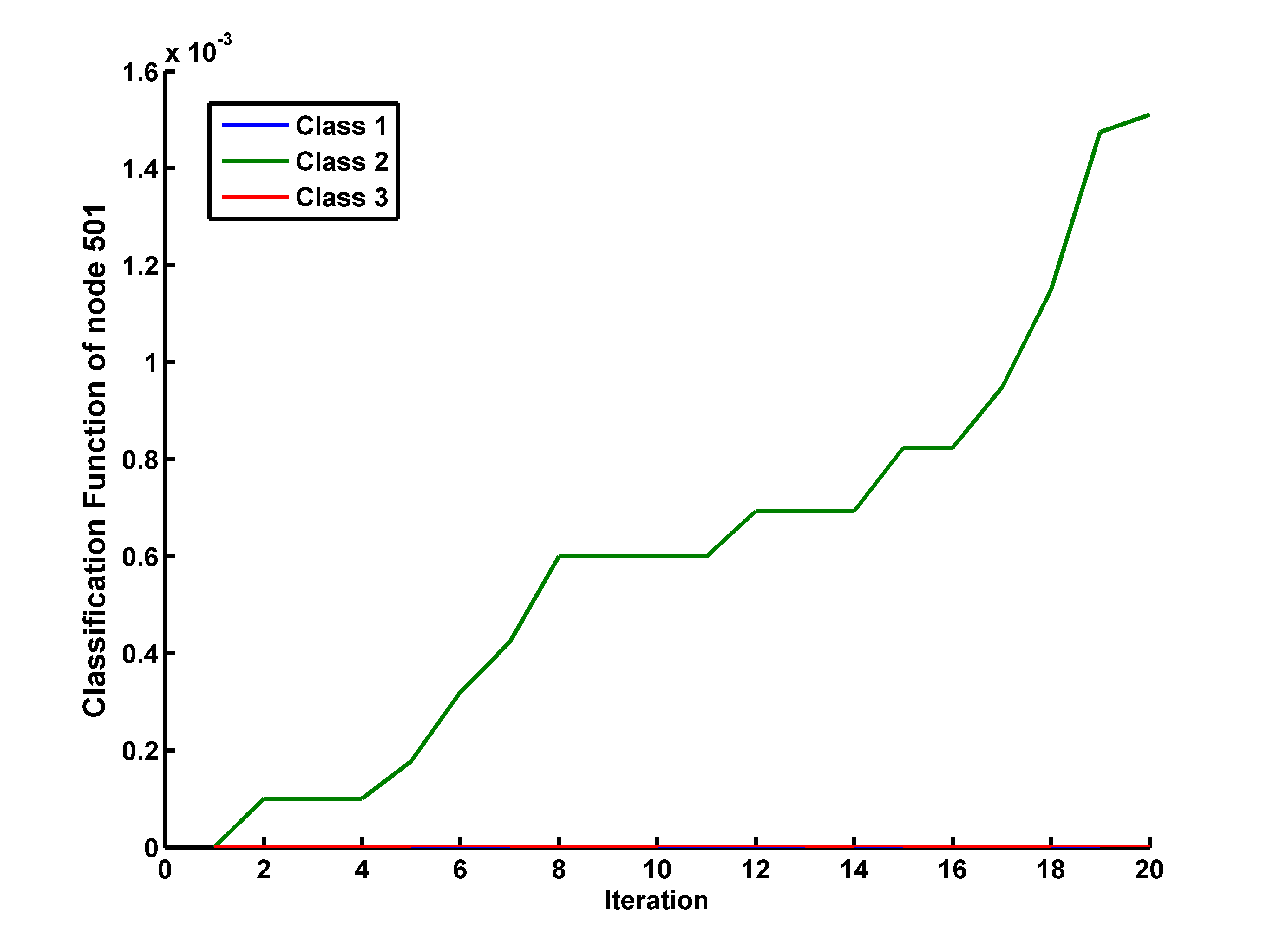}
\caption{The figure shows how the value of the feature vector elements corresponding to each class changes as the iterations increase. The new node that came in belonged to class $2$. The feature value for class $1$ and class $3$ does not increase because the node does not have any neighbours belonging to any of those classes.}
\label{11}
\end{center}
\end{figure}

The tracking ability of the sampling algorithm is demonstrated next. After the sampling algorithm ran for $200$ iterations on the original graph with $500$ nodes, a new node was introduced using the same parameters into the graph. The algorithm continued from the classification function found after $200$ iterations for other $20$ iterations, this time with $501$ nodes. The class of all the nodes was recomputed. The evolution of the feature vector of the new node is shown in \figurename{11}. In the example presented here, the new node belonged to class $2$. It can be seen from \figurename{11} that the classification function value for the class $2$ increases as the number of iterations increase and hence the new node gets classified correctly, i.e., it has been correctly tracked.

\begin{figure}[h]
\begin{center}
\includegraphics[scale=0.08]{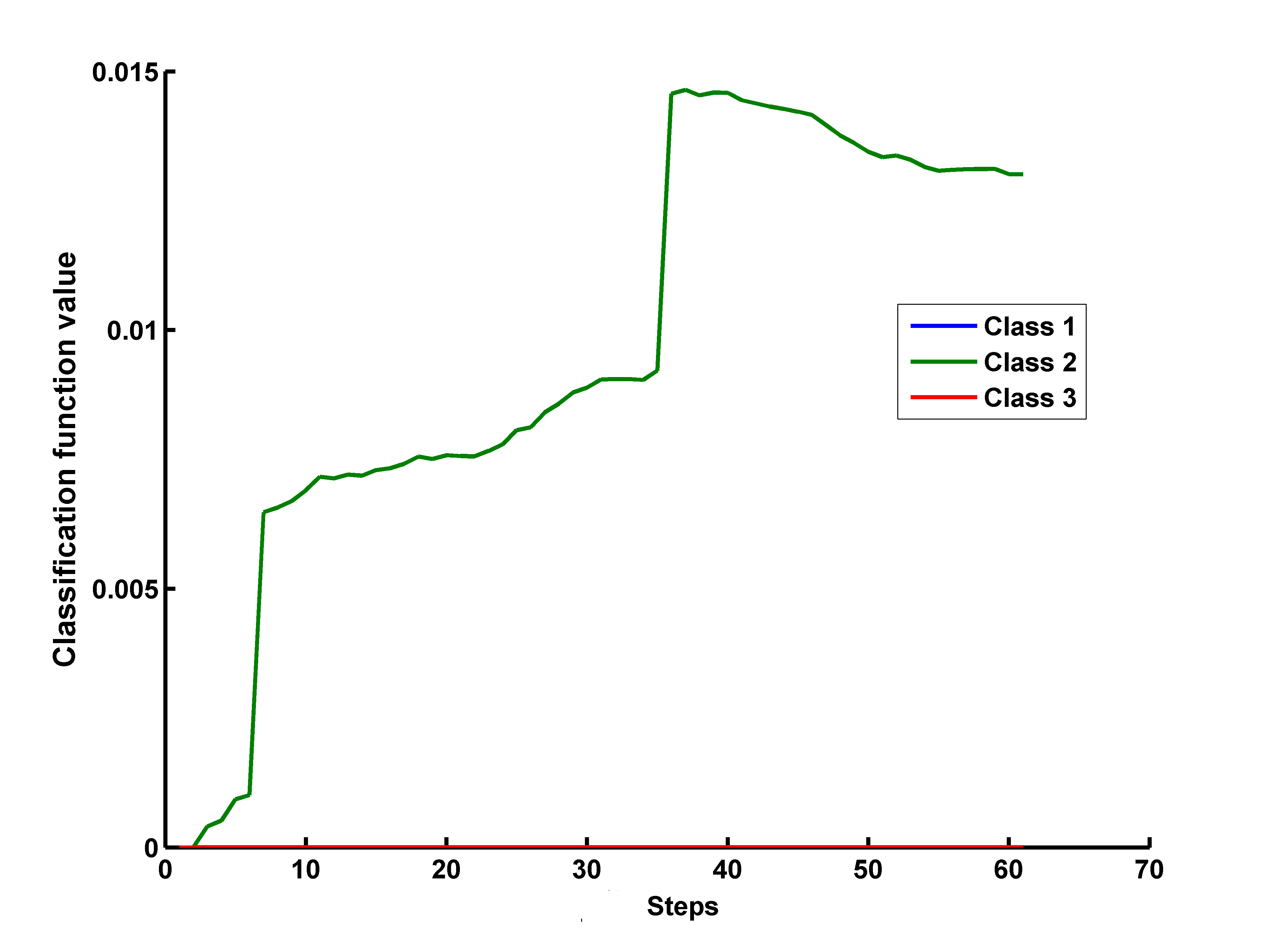}
\caption{Classification function v/s number of steps of the selection method for an alternate node selection technique.}
\label{12}
\end{center}
\end{figure}

Instead of the MCMC based sampling, the class of the new incoming node can also be found by using an alternate node selection scheme. In this, we first select the incoming node and sample from its neighbors. After that, we select the neighbors of the incoming node in a round robin fashion and sample from their neighbors. Once all the 1-neighbors of the incoming node have been exhausted, we again select the incoming node and sample from its neighbors. The evolution of classification function of the new incoming node for such a node selection technique has been shown in \figurename{12}. The node is classified to its correct class by this technique as well. The idea behind its working is that an unlabelled incoming node will not change the classification function of the other nodes very much. In fact, it will hardly change the values for nodes that are multiple hops away. So selecting nodes as described in this experiment is sufficient to get a good estimate of the classification function. Indeed, this technique is much faster than the MCMC technique because we do not select nodes from the entire graph, but only from 1-hop neighbors.

\subsection{Stochastic Block Model}

We now consider a dynamic graph in which nodes can enter and leave the graph. It has been demonstrated that the sampling algorithm can classify new incoming nodes. A dynamic Stochastic Block Model with multiple classes is taken.
Stochastic block models \cite{CK01,HLM12} have been found to be closer to reality than classical models such as preferential attachment, in applications such as social networks and there is a growing interest in them. Intuitively, they have relatively denser subnetworks which are interconnected to each other through weak ties.
A node is connected with another node of the same class with probability $p_{in}$ and node from a different class with probability $p_{out}$; $p_{in} >> p_{out}$.
In the present work we introduce an extension of the Stochastic Block Model to the dynamic setting. Specifically, there is a Poisson arrival of nodes into the graph with rate $\lambda_{arr}$. The class of the arriving node is chosen according to a pre-specified probability distribution. Each node stays in the graph for a random time that is exponentially distributed with mean $\mu_{dep}$ after which it leaves. The maximum number of nodes in the graph is limited to $K$, i.e., nodes arriving when the number of nodes in the graph is $K$ do not become a part of the graph. This system can be modelled as a M/M/K/K queue. As a result, irrespective of the number of nodes that the graph has initially, the average the number of nodes in the graph will reach a steady state value given by $\frac{\lambda_{arr}}{\mu_{dep}}$.

\begin{figure}[h]
\centering
\begin{subfigure}[a]{0.325\linewidth}
\centering
\includegraphics[width=\linewidth]{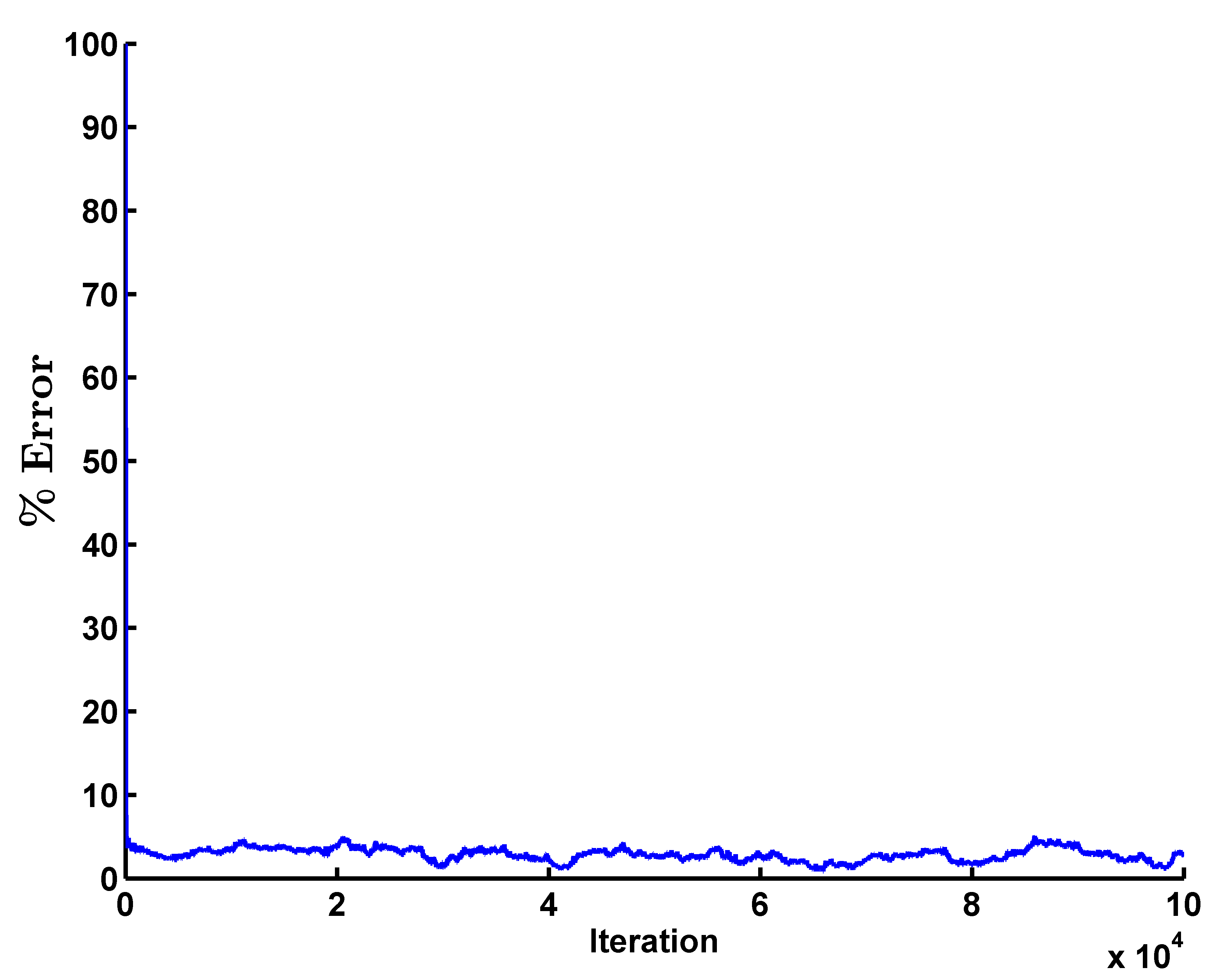}
\captionof{figure}{$\%$error v/s iterations}
\end{subfigure}
\hfill
\begin{subfigure}[a]{0.325\linewidth}
\centering
\includegraphics[width=\linewidth]{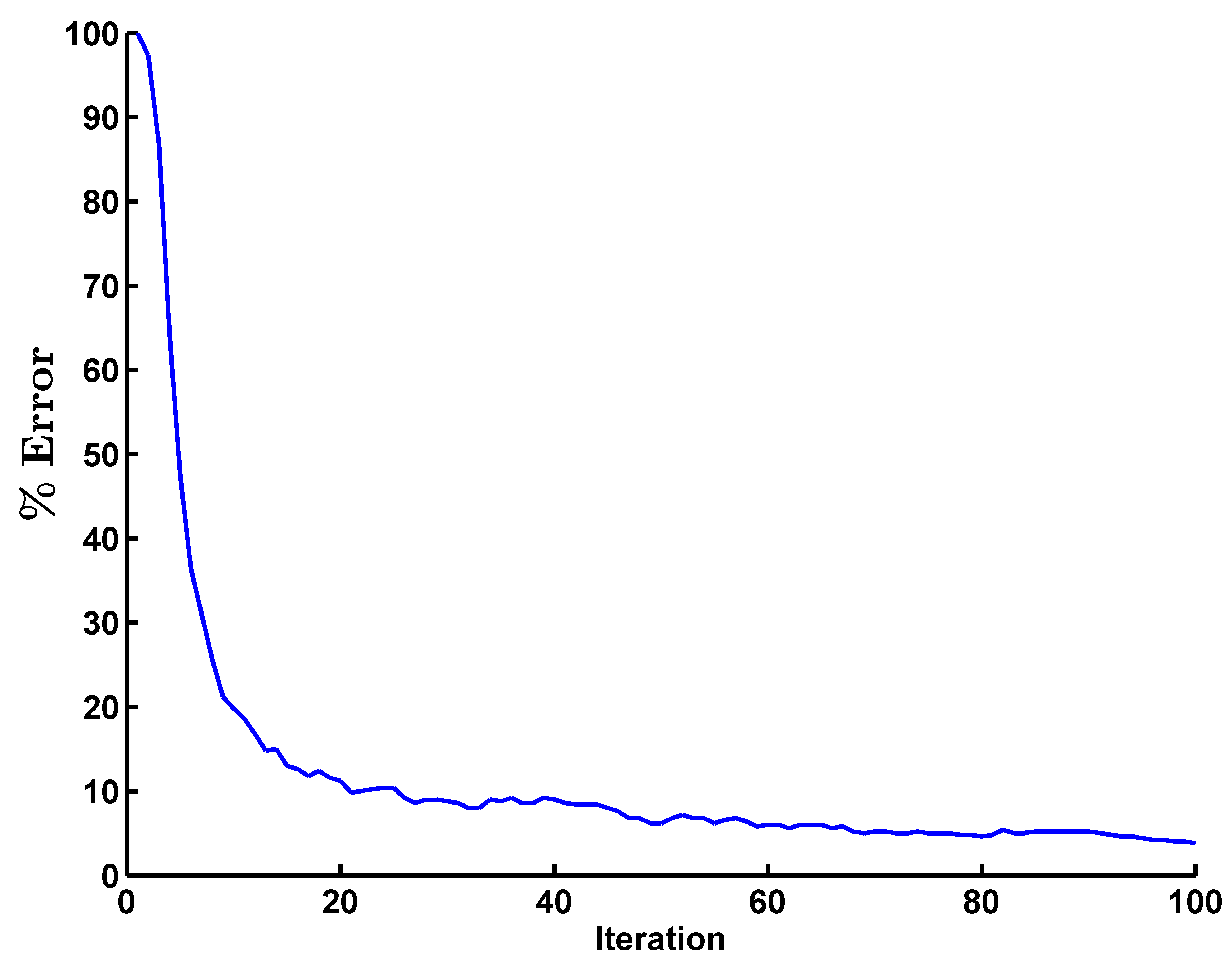}
\captionof{figure}{First 100 iterations}
\end{subfigure}
\hfill
\begin{subfigure}[a]{0.325\linewidth}
\centering
\includegraphics[width=\linewidth]{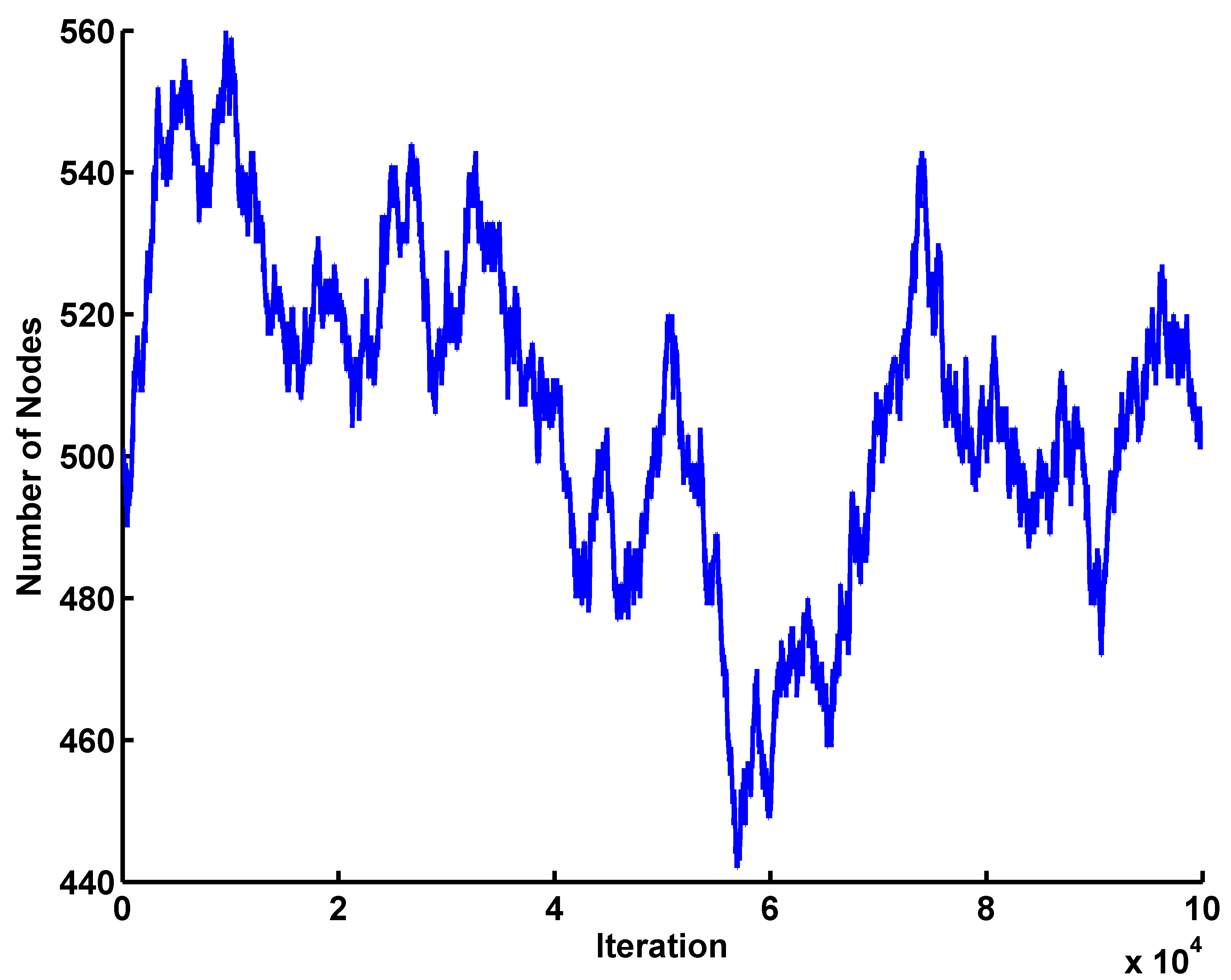}
\captionof{figure}{Size of graph v/s iterations}
\end{subfigure}
\caption{500 nodes during initialisation}
\label{13}
\end{figure}

\begin{figure}[h]
\centering
\begin{subfigure}[a]{0.325\linewidth}
\centering
\includegraphics[width=\linewidth]{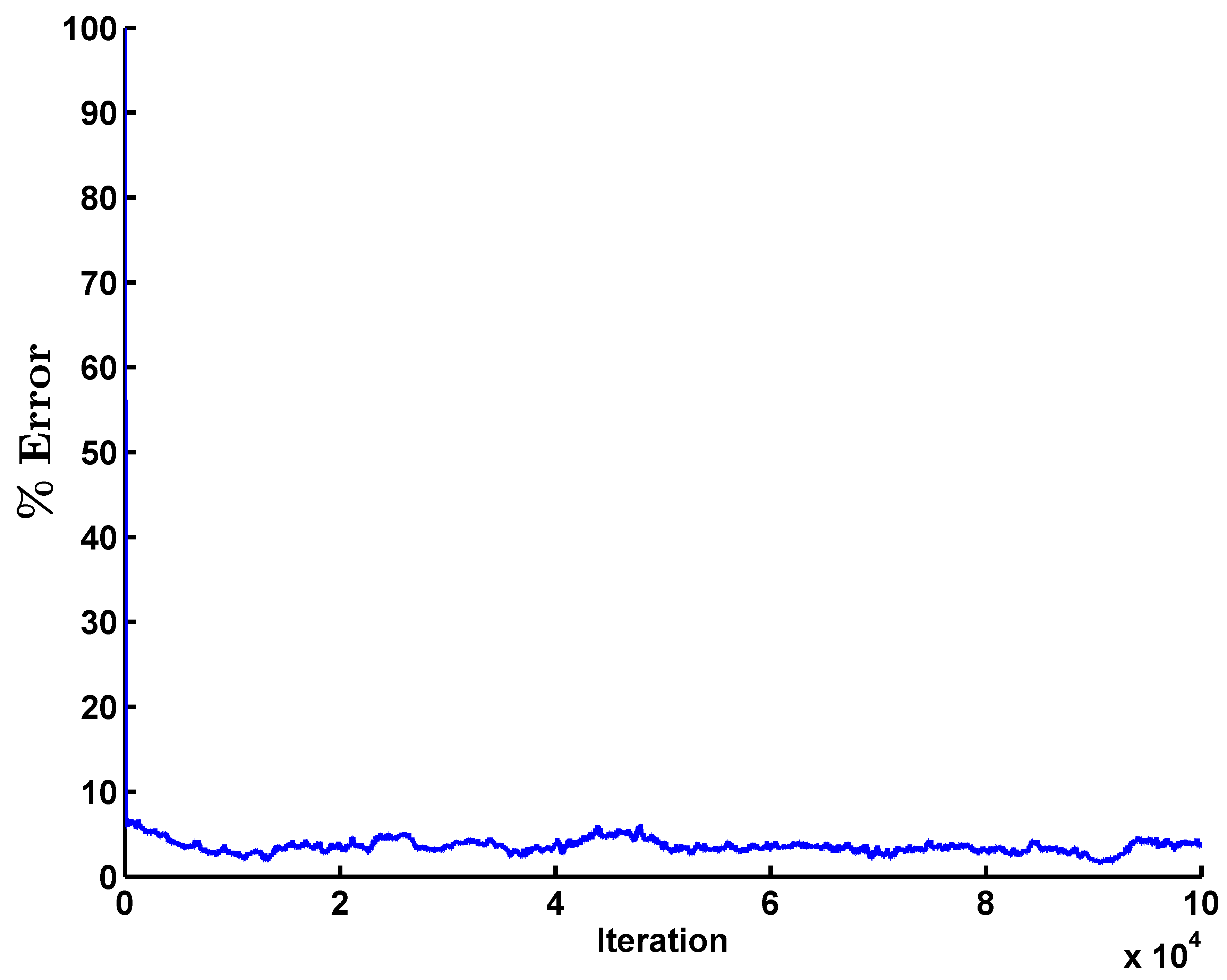}
\captionof{figure}{$\%$error v/s iterations}
\end{subfigure}
\hfill
\begin{subfigure}[a]{0.325\linewidth}
\centering
\includegraphics[width=\linewidth]{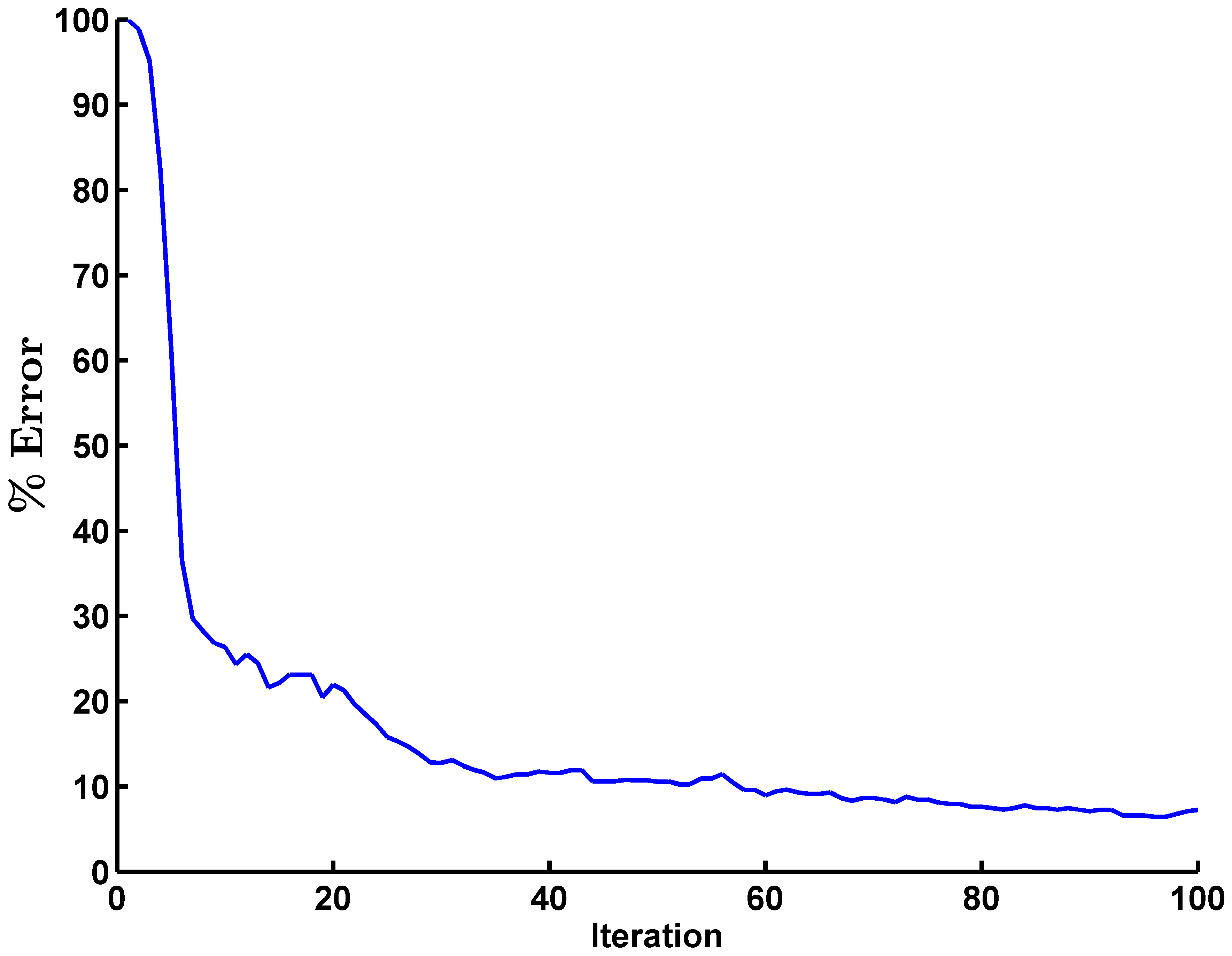}
\captionof{figure}{First 100 iterations}
\end{subfigure}
\hfill
\begin{subfigure}[a]{0.325\linewidth}
\centering
\includegraphics[width=\linewidth]{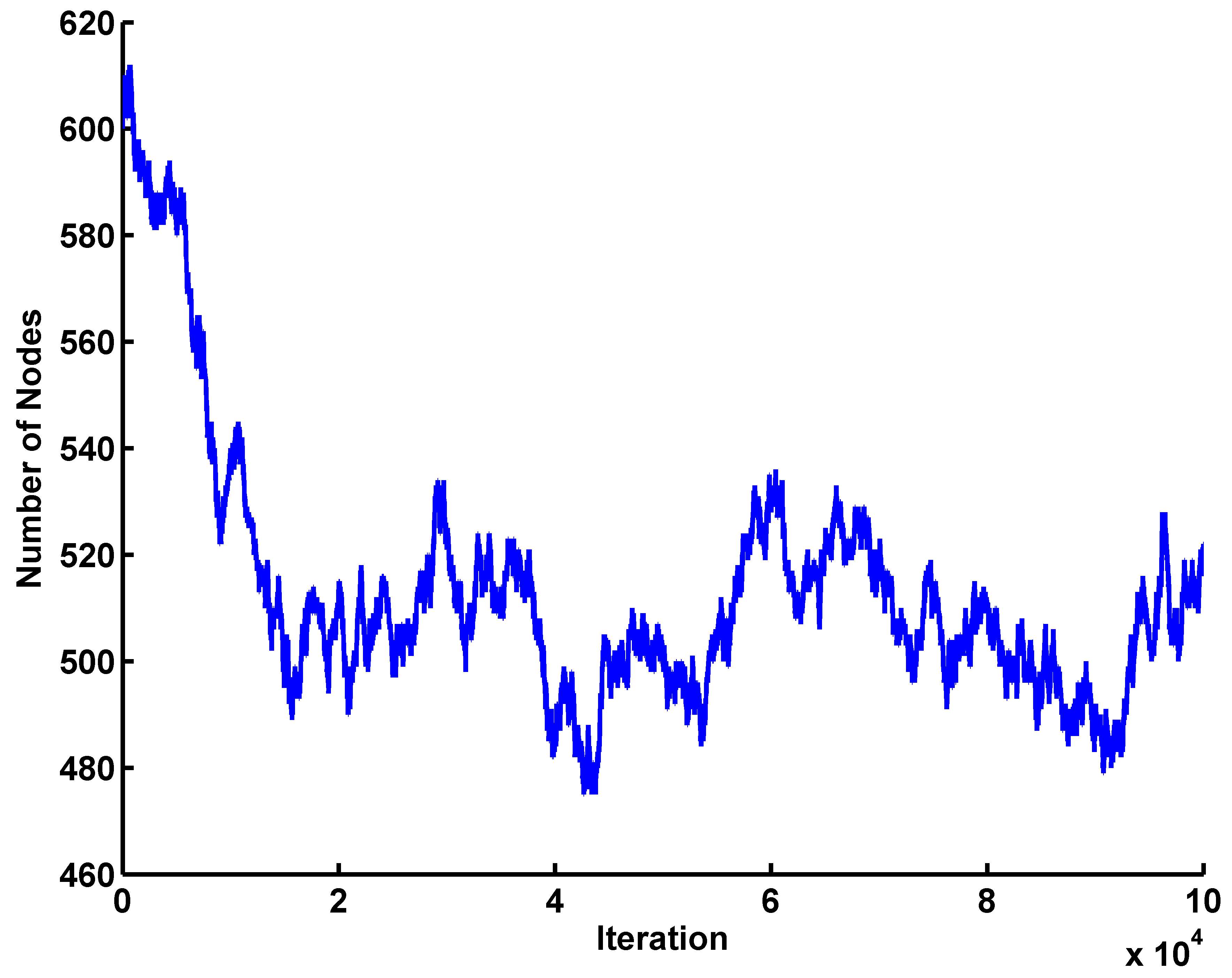}
\captionof{figure}{Size of graph v/s iterations}
\end{subfigure}
\caption{600 nodes during initialisation}
\label{14}
\end{figure}

\begin{figure}[h]
\centering
\begin{subfigure}[a]{0.325\linewidth}
\centering
\includegraphics[width=\linewidth]{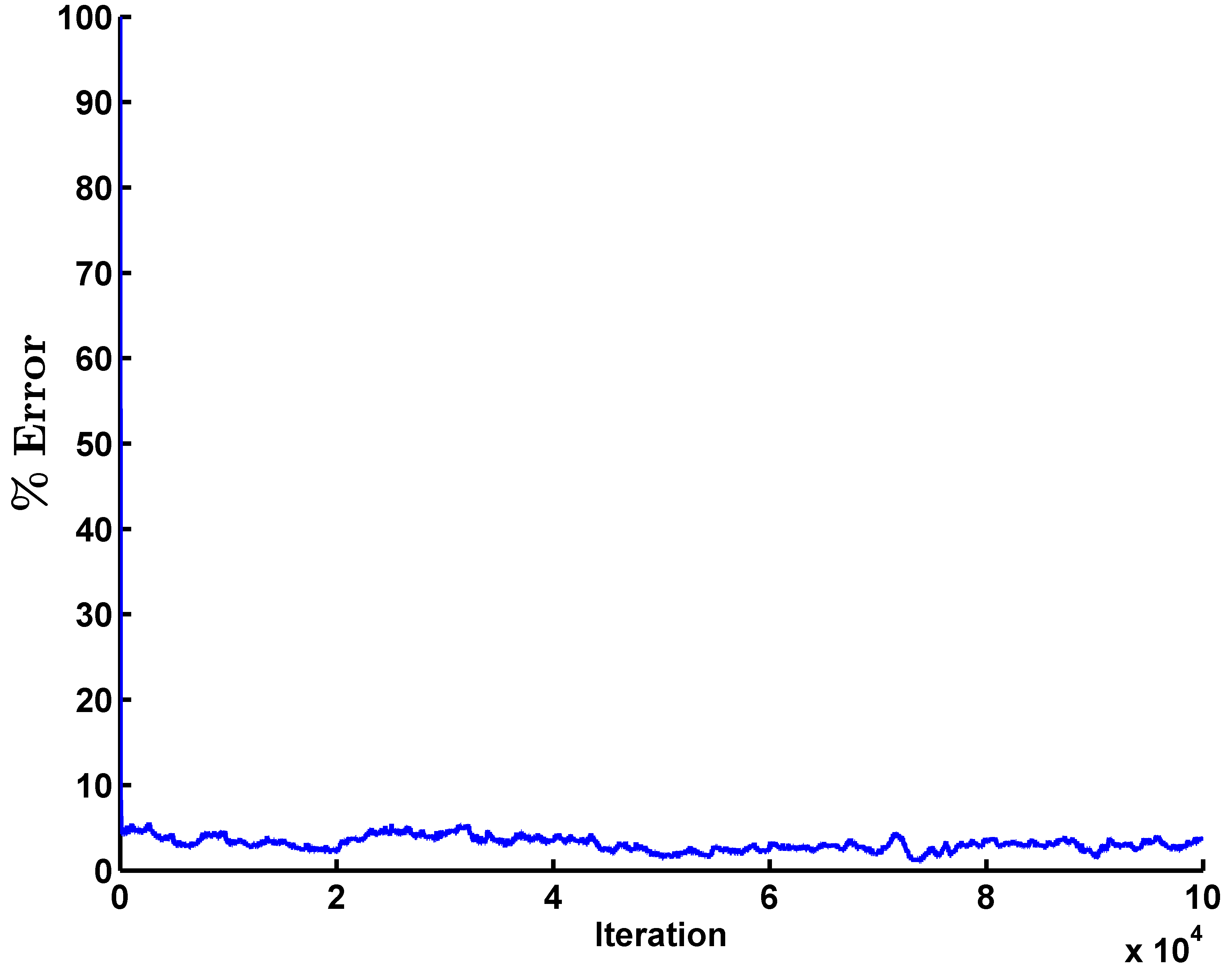}
\captionof{figure}{$\%$error v/s iterations}
\end{subfigure}
\hfill
\begin{subfigure}[a]{0.325\linewidth}
\centering
\includegraphics[width=\linewidth]{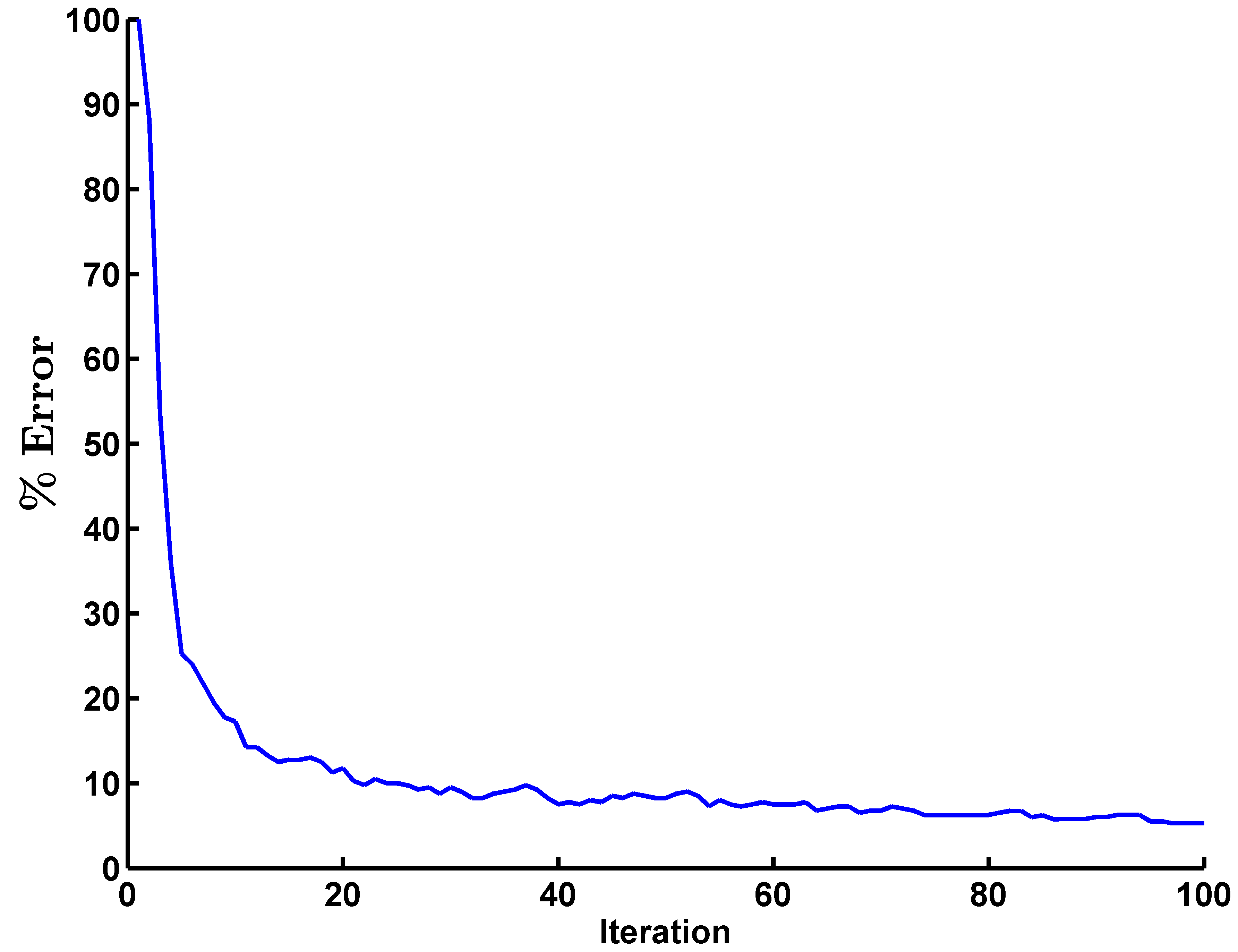}
\captionof{figure}{First 100 iterations}
\end{subfigure}
\hfill
\begin{subfigure}[a]{0.325\linewidth}
\centering
\includegraphics[width=\linewidth]{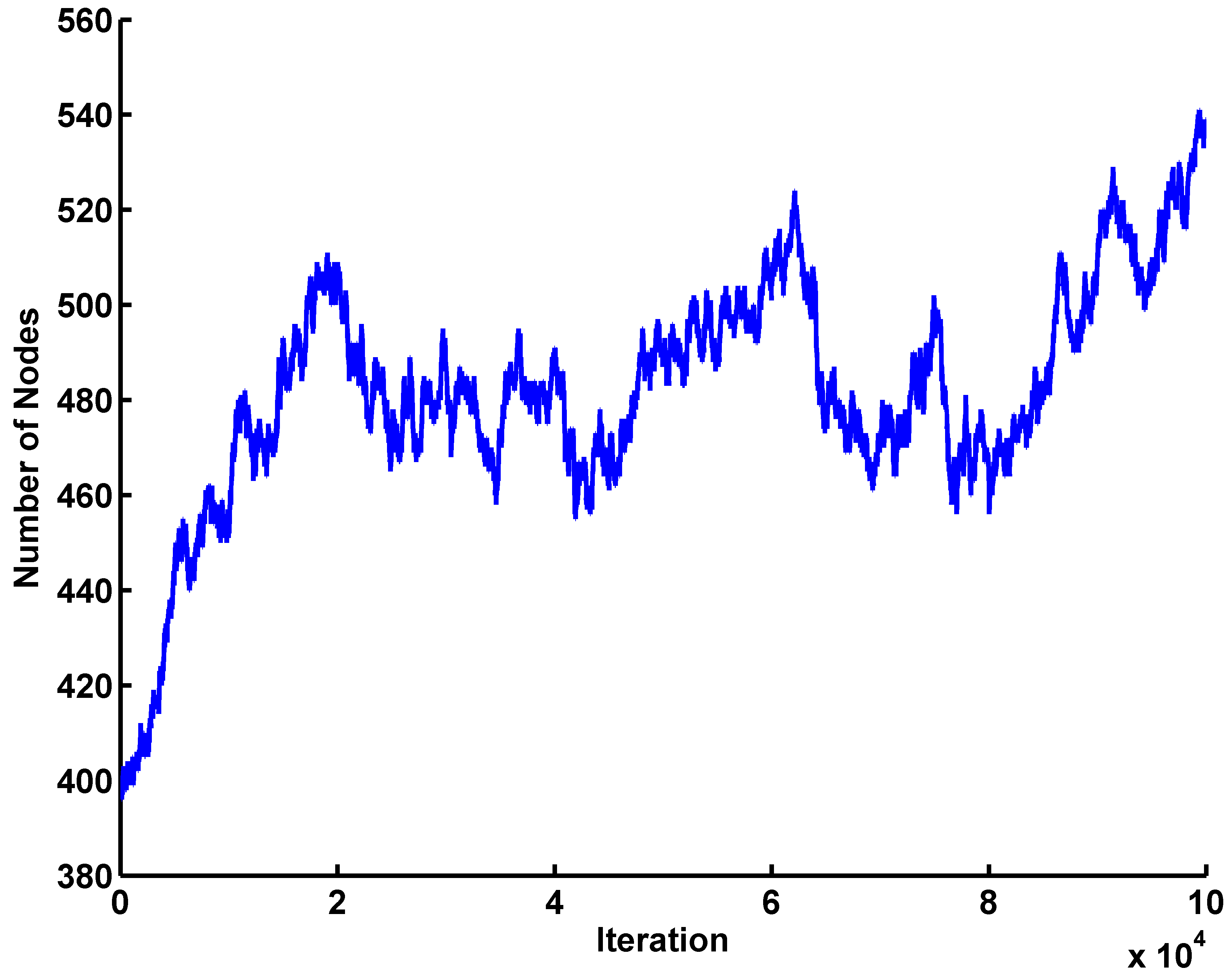}
\captionof{figure}{Size of graph v/s iterations}
\end{subfigure}
\caption{400 nodes during initialisation}
\label{15}
\end{figure}

Simulations were performed for various values of $\lambda_{arr}$ and $\mu_{dep}$ with different number of nodes during initialisation. The graph had $3$ classes and an incoming node could belong to either of the classes with equal probability. $K=1000$, $\lambda_{arr} = 1/(2\times10^4)$ and $\mu_{dep} = 1/10^7$ was chosen. Therefore, $\frac{\lambda_{arr}}{\mu_{dep}} = 500$. Two nodes with the maximum degree from each class were chosen as the labelled nodes during initialization. $K$ steps were considered as one iteration. \figurename{13}, \figurename{14} and \figurename{15} show the evolution of $\%$error and the size of the graph for initial graph size of $500$, $600$ and $400$ respectively. The step size for these simulations was kept constant at $1/1000$ so that the algorithm does not lose its tracking ability. This is common practice in tracking applications. Since a constant stepsize $\eta_t \equiv \eta > 0$ does not satisfy the condition $\sum_t\eta_t^2 < \infty$, one cannot claim convergence with probability one even in the stationary environment. Instead, one has the weaker claim that the asymptotic probability of the iterates not being in a small neighborhood of the desired limit is small, i.e., $O(\eta)$ (\cite{book}, Chapter 9). The problem with using decreasing stepsize in time-varying environment can be explained as follows. With the interpretation of stochastic approximation as a noisy discretization of the ODE, one views the step-size as a time step, whence it dictates the time scale on which the algorithm moves. A decreasing step-size schedule begins with high step-sizes and slowly decreases it to zero, thus beginning with a fast time scale for rapid exploration at the expense of larger errors, ending with slower and slower time scales that exploit the problem structure ensuring more better error suppression and graceful convergence.  For tracking applications, the environment is non-stationary, thus the decreasing step-size scheme loses its tracking ability once the algorithmic time scale becomes slower than that of the environment. A judiciously chosen constant step-size avoids this problem at the expense of some accuracy. In our algorithms, the Perron-Frobenius eigenvalue $\lambda$ of the matrix $\alpha B$ is the key factor dictating the convergence rate, hence the tracking scheme will work well if $\lambda_{arr}, \mu_{dep} << \lambda$.

The number of misclassified nodes is higher in the stochastic block model (3-7$\%$) than in the Gaussian mixture model. Most of these errors are because node of one class is connected to the labelled node of some other class. Error in some other nodes is because the classification function value is very close for $2$ out of the $3$ classes. It was also observed that on increasing  cluster density, the average error decreases in steady state since some nodes which got wrongly classified by being close to an influential node in some other cluster now will have more neighbors of the same class.

\begin{figure}[h]
\centering
\begin{subfigure}[a]{0.49\linewidth}
\centering
\includegraphics[width=0.7\linewidth]{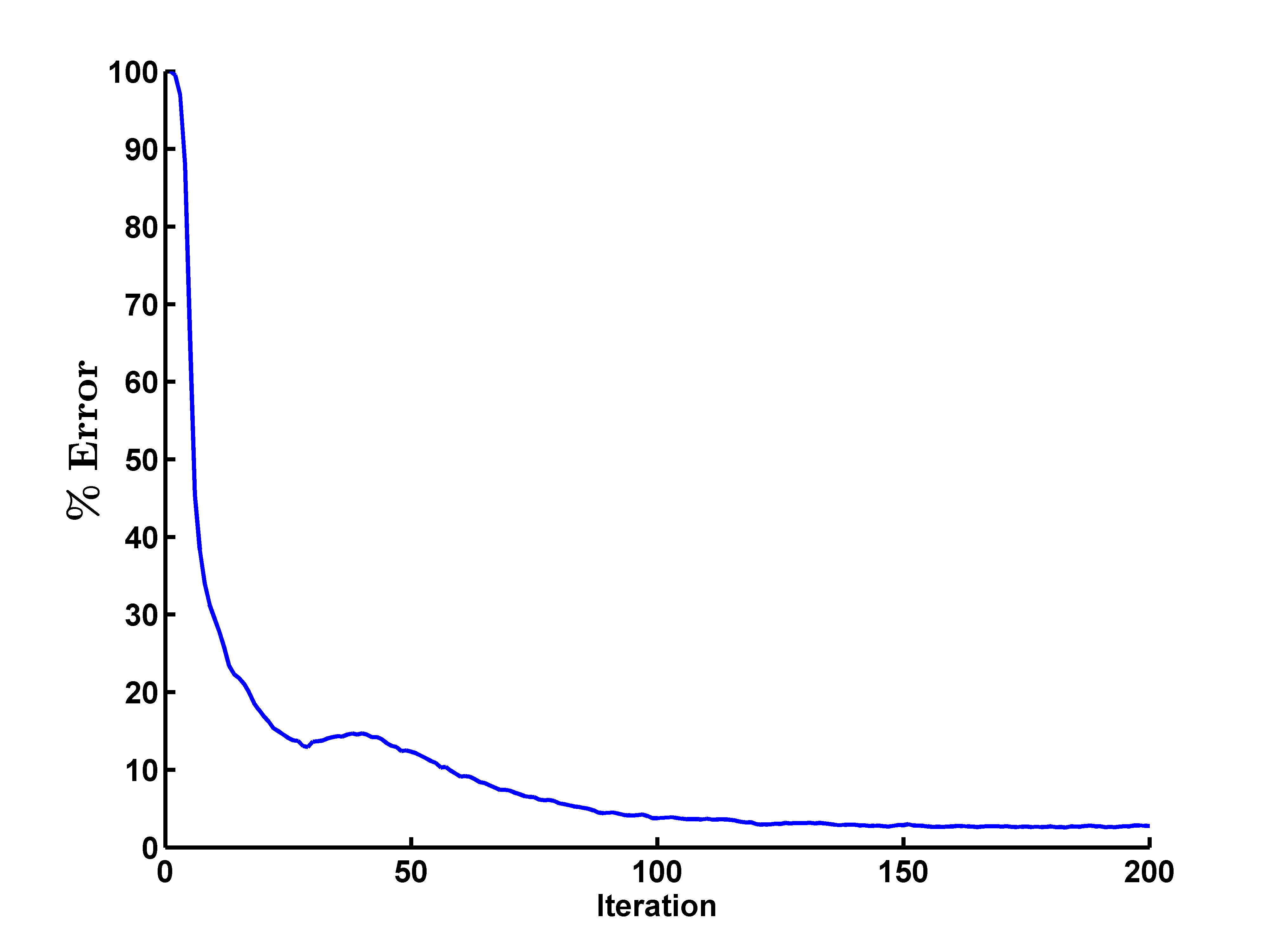}
\captionof{figure}{$\%$error v/s iterations}
\end{subfigure}
\begin{subfigure}[a]{0.49\linewidth}
\centering
\includegraphics[width=0.7\linewidth]{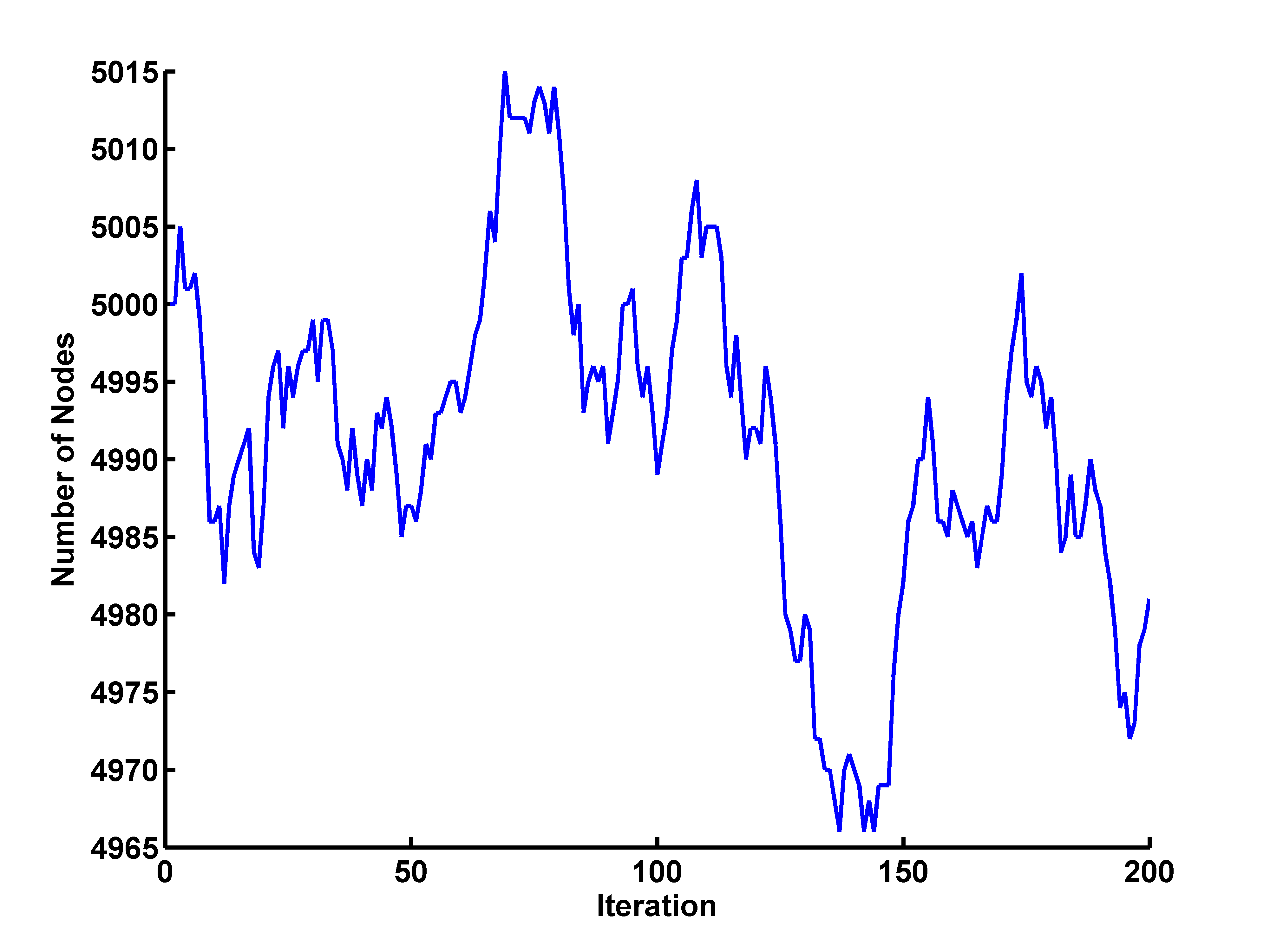}
\captionof{figure}{Size of graph v/s iterations}
\end{subfigure}
\caption{Increased arrival rate with 5000 nodes during initialisation}
\label{16}
\end{figure}

\begin{figure}[h]
\centering
\begin{subfigure}[a]{0.49\linewidth}
\centering
\includegraphics[width=0.7\linewidth]{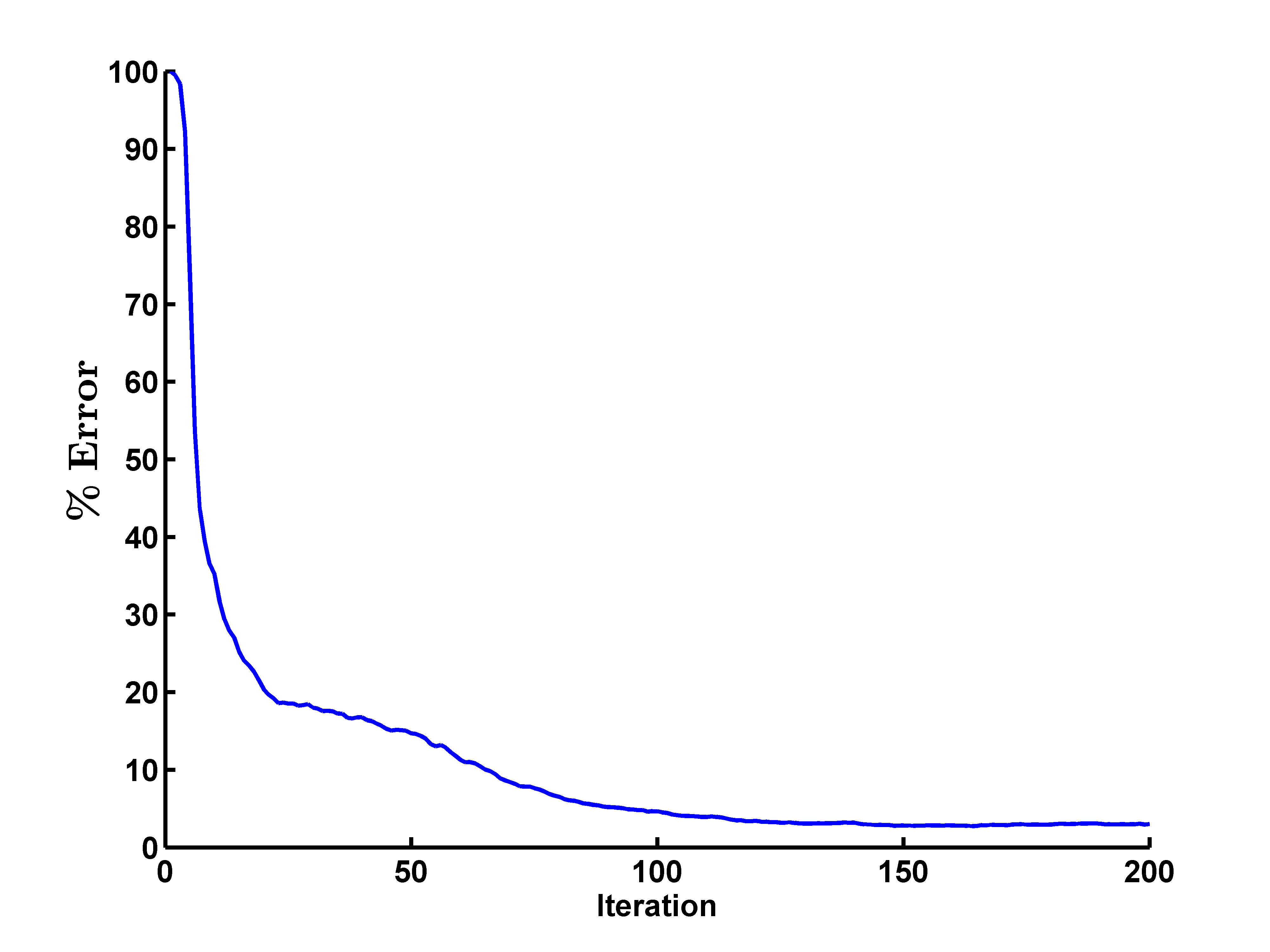}
\captionof{figure}{$\%$error v/s iterations}
\end{subfigure}
\begin{subfigure}[a]{0.49\linewidth}
\centering
\includegraphics[width=0.7\linewidth]{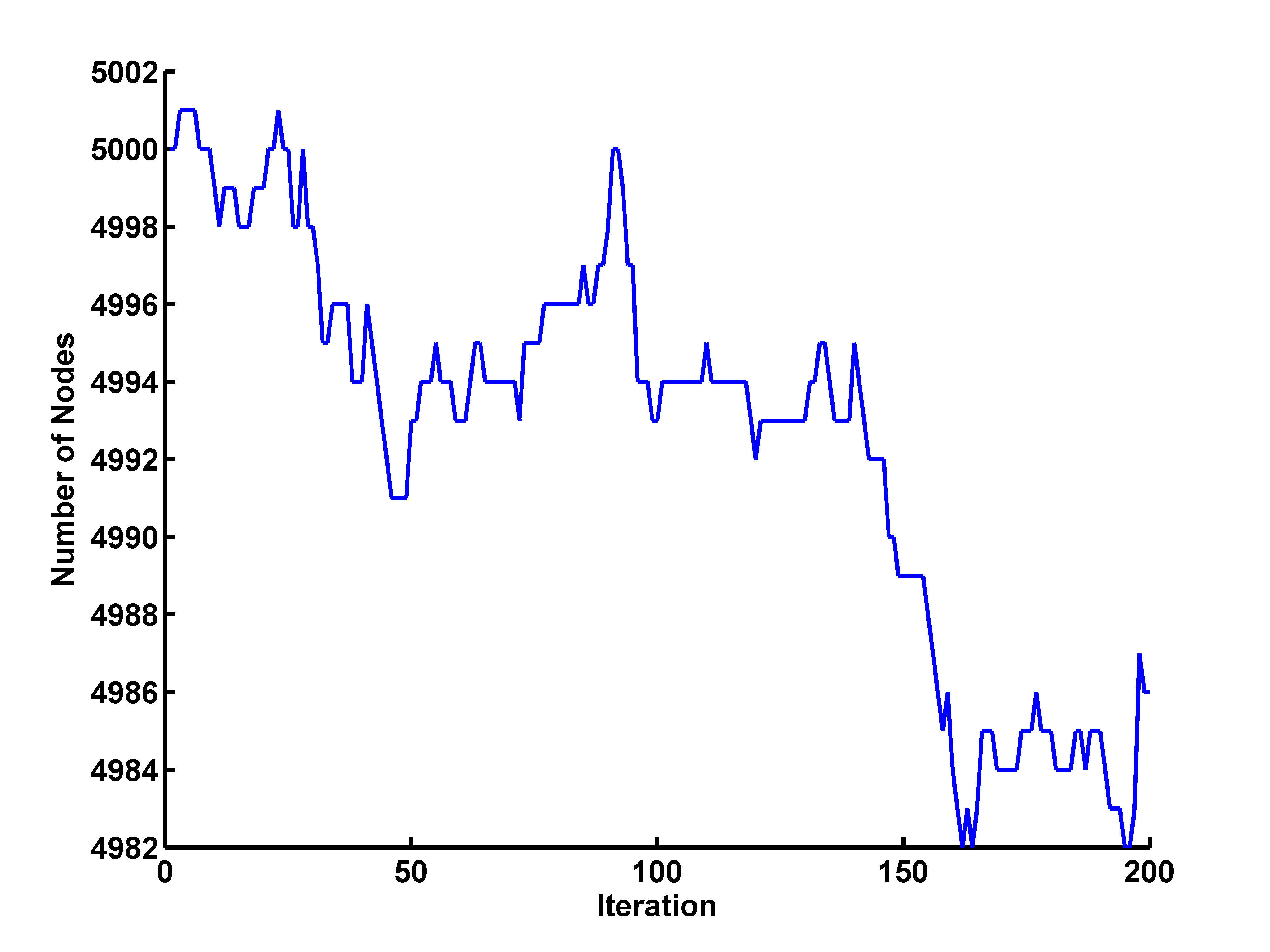}
\captionof{figure}{Size of graph v/s iterations}
\end{subfigure}
\caption{Decreased departure rate with 5000 nodes during initialisation}
\label{17}
\end{figure}

\figurename{16} shows the effect of increasing the arrival rate to $\lambda_{arr} = 1/(2\times 10^3)$ while keeping $\mu_{dep}$ the same. Similarly, \figurename{17} shows the effect of decreasing the departure rate to $\mu_{dep} = 1/10^8$ while keeping $\lambda_{arr} = 1/(2\times 10^4)$. For both these cases, the number of nodes in steady state should be 5000. It can be seen that the decrease in error is much more smoother in this case as compared to \figurename{13}(b),\figurename{14}(b) and \figurename{15}(b) which is indicative of the slowly varying size of the graph.

During the experiment, the number of labelled nodes from each cluster were kept constant. In case a labelled node left, then a neighbour of the leaving node belonging to the same cluster was randomly chosen as the labelled node. If this wasn't done, the misclassification error became as high as $70\%$ in some cases. This can be reasoned as being due to the absence of labelled nodes resulting in increase in cluster size of other clusters. A similar effect was observed if the number of labelled nodes in one cluster was much more compared to the others.

\begin{figure}[h]
\centering
\begin{subfigure}[a]{0.325\linewidth}
\centering
\includegraphics[width=\linewidth]{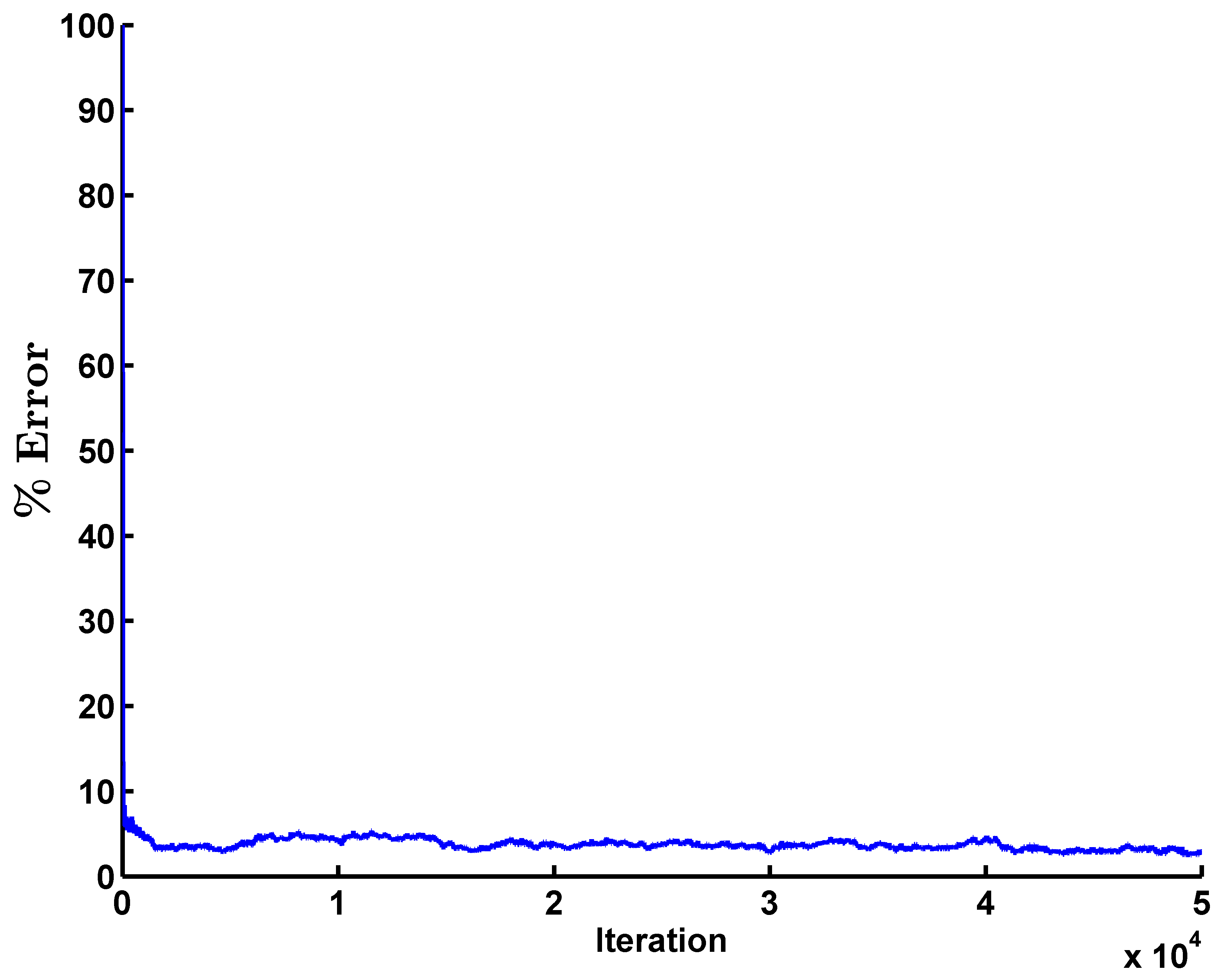}
\captionof{figure}{$\%$error v/s iterations}
\end{subfigure}
\hfill
\begin{subfigure}[a]{0.325\linewidth}
\centering
\includegraphics[width=\linewidth]{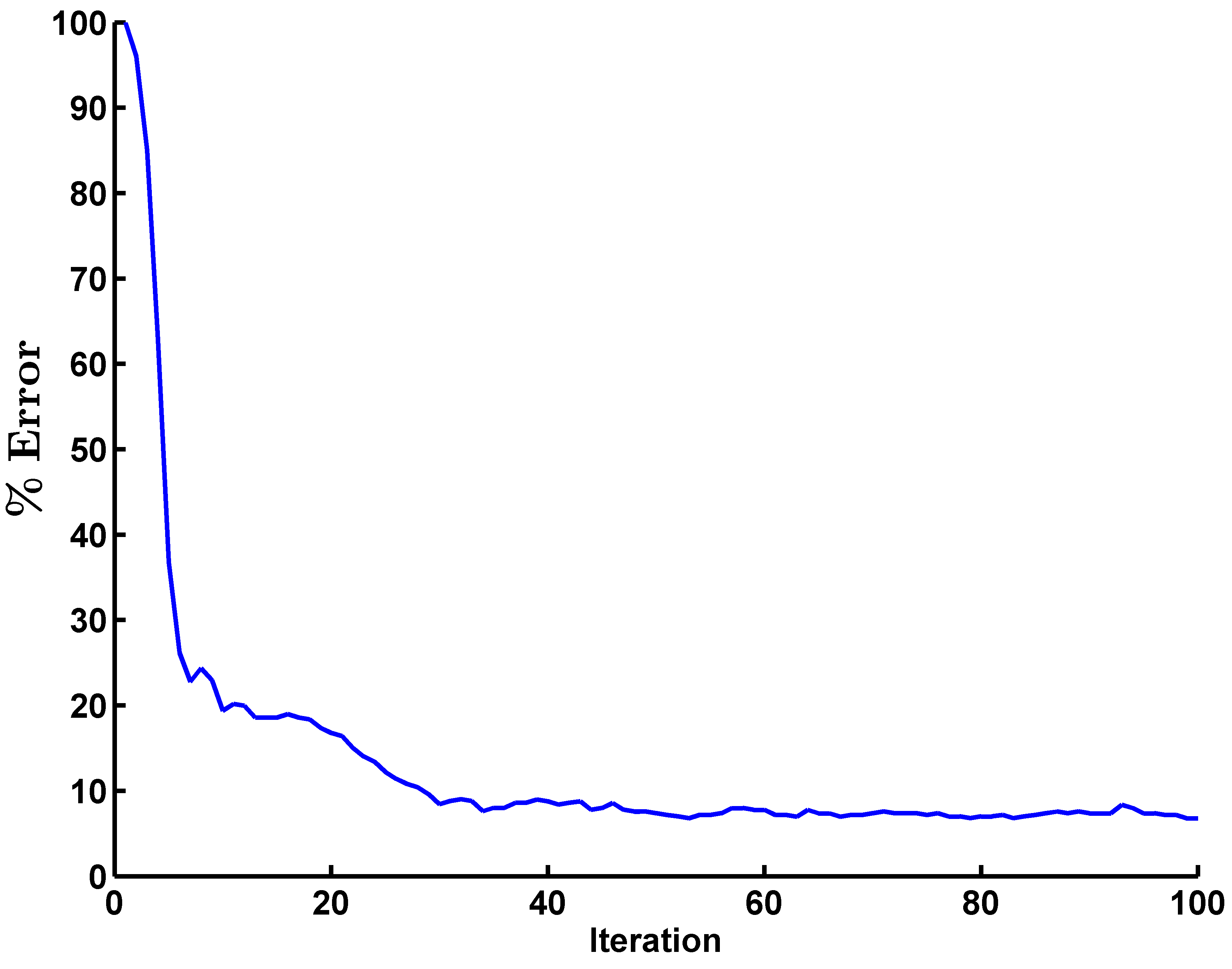}
\captionof{figure}{First 100 iterations}
\end{subfigure}
\hfill
\begin{subfigure}[a]{0.325\linewidth}
\centering
\includegraphics[width=\linewidth]{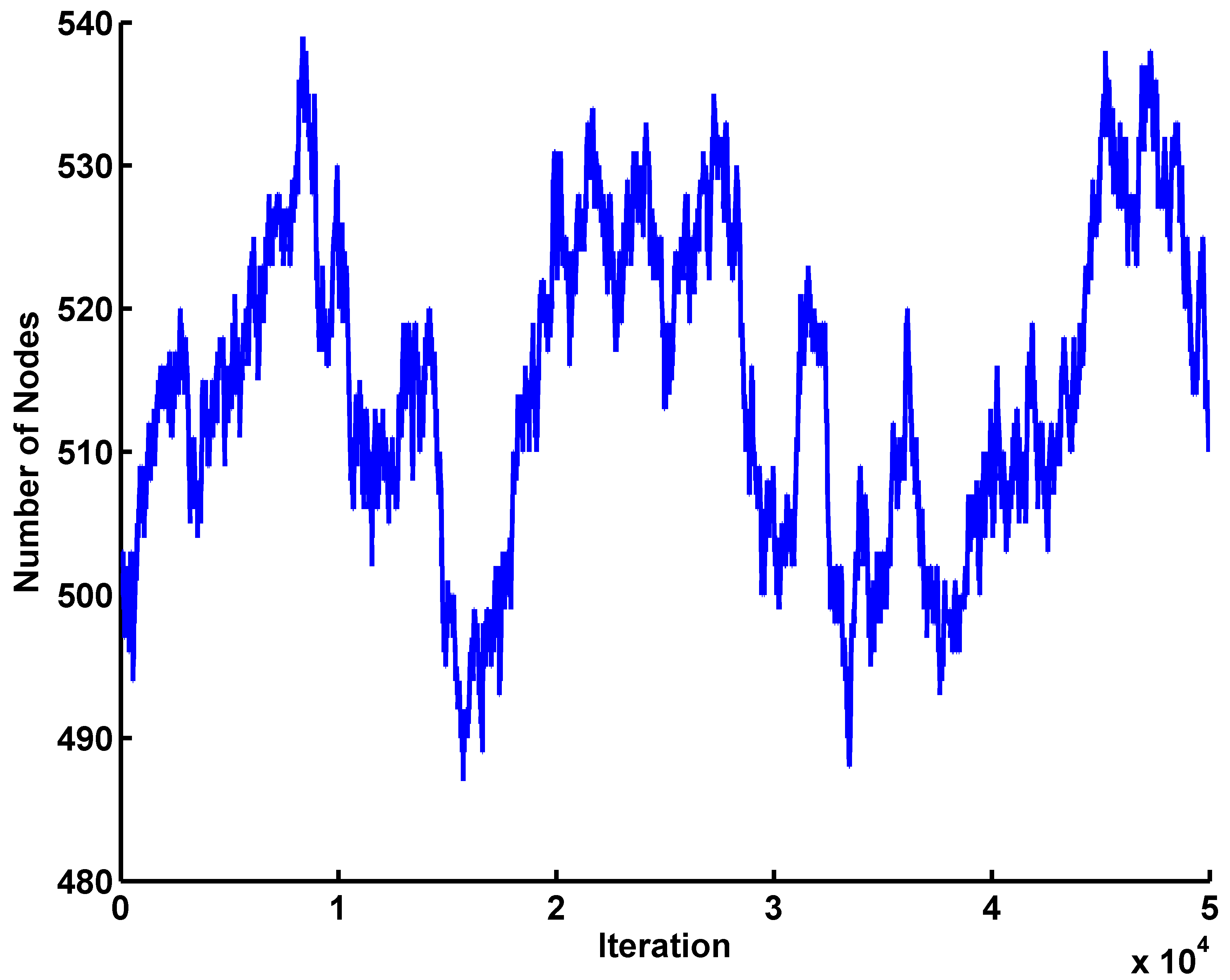}
\captionof{figure}{Size of graph v/s iterations}
\end{subfigure}
\caption{Permanently labelled nodes with 500 nodes during initialisation}
\label{18}
\end{figure}

To better understand how the departure of a labelled node effects the error, we performed tests with graphs in which the labelled nodes were permanent, i.e., the labelled nodes did not leave. \figurename{18} shows the variation of error for this case. It can be seen that in this case, the variance in error after reaching steady state is much less as compared to \figurename{13}(a), \figurename{14}(a) and \figurename{15}(a).

\section{Conclusion}
\label{sec5}
Two iterative algorithms for graph based semi-supervised learning have been proposed. The first algorithm is based on  iteration of an affine contraction (w.r.t,.\ a weighted norm) while the second algorithm is based on a sampling scheme inspired by reinforcement learning. The classification accuracy of the nodes of the graph into classes for two algorithms was evaluated and confirmed to be the same. The performance of the sampling algorithm was also evaluated on Gaussian mixture graphs and it was found that the nodes other than the boundary nodes are classified correctly. The ability to track the feature vector of a new node that arrives during the simulation was tested. The sampling algorithm correctly classifies the new incoming node in considerably fewer iterations. This ability was then applied to a dynamic stochastic block model graph modelled based on M/M/K/K queue. It was demonstrated that the sampling algorithm can be applied to dynamically changing systems and still achieve a small error. The sampling algorithm can be implemented in a distributed manner unlike the power iteration algorithm and has very little per iterate computation, hence it should be the preferred methodology for very large graphs.\\



\begin{thebibliography}{}

\bibitem{Aetal08}
Avrachenkov, K., Dobrynin, V., Nemirovsky, D., Pham, S. K., and Smirnova, E.
``Pagerank based clustering of hypertext document collections''.
In Proceedings of ACM SIGIR 2008, pp.873-874, 2008.

\bibitem{semisup}
Avrachenkov, K., Gon\c{c}alves, P., Mishenin, A., and Sokol, M.  ``Generalized optimization framework for graph-based semi-supervised learning.'' \textit{Proceedings of SIAM Conference on Data Mining (SDM 2012)}. Vol. 9. 2012.

\bibitem{AGS13}
Avrachenkov, K., Gon\c{c}alves, P., and Sokol, M.
On the choice of kernel and labelled data in semi-supervised learning methods.
In Proceedings of Algorithms and Models for the Web Graph (WAW 2013), pp.56-67, 2013.

\bibitem{BG09}
Bell N., Garland M.
``Implementing sparse matrix-vector multiplication on throughput-oriented processors''.
In Proceedings of the ACM Conference on High Performance Computing Networking, Storage and Analysis. 2009.

\bibitem{reinf}
Borkar, V.S., and Mathkar, A.S. ``Reinforcement Learning for Matrix Computations: PageRank as an Example.'' \textit{Distributed Computing and Internet Technology, Proceedings of the 10th ICDCIT, Bhubaneshwar, India} (R.\ Natarajan, ed.) Springer Lecture Notes in Computer Science No.\ 8337,  Springer International Publishing, Switzerland, 2014. 14-24.

\bibitem{book}
Borkar, V.S. \textit{Stochastic approximation: A Dynamical Systems Viewpoint}. Hindustan Publishing Agency, New Delhi, and Cambridge Uni.\ Press, Cambridge, UK, 2008.

\bibitem{CSZ06}
Chapelle, O., Sch\"{o}lkopf, B. and Zien A.
{\it Semi-supervised learning}, MIT Press, 2006.

\bibitem{CK01}
Condon, A., and Karp, R. ``Algorithms for graph partitioning on the planted partition model''.
{\it Random Structures and Algorithms}, v.18, pp.116-140, 2001.

\bibitem{FoussPirotte06Ker}
Fouss, F., Yen, L., Pirotte, A., and Saerens, M.
``An experimental investigation of graph kernels on a collaborative recommendation task''.
In \emph{Sixth International Conference on Data Mining (ICDM'06)}, pp.863-868, 2006.

\bibitem{FoussFrancoisseSaerens11}
Fouss, F., Francoisse, K., Yen, L., Pirotte A., and Saerens, M.
``An experimental investigation of kernels on graphs for collaborative recommendation and semisupervised classification''.
\emph{Neural Networks}, 31, pp.53-72, 2012.

\bibitem{lesmis}
Knuth, D.E. The Stanford GraphBase: A Platform for Combinatorial Computing, Addison-Wesley, Reading, MA 1993.

\bibitem{HLM12}
Heimlicher, S., Lelarge, M., and Massouli\'e, L. ``Community detection in the labelled stochastic
block model''. ArXiv preprint arXiv:1209.2910, 2012.

\bibitem{Netal08}
John Nickolls, J., Ian Buck, I., Michael Garland, M., and Kevin Skadron, K.
``Scalable parallel programming with CUDA''. {\it Queue}, v.6(2), pp.40-53, 2008.

\bibitem{SAT10}
Smirnova, E., Avrachenkov, K., and Trousse, B.
``Using Web Graph Structure for Person Name Disambiguation''.
In Proceedings of CLEF, v.77, 2010.

\bibitem{Zetal04}
Zhou, D., Bousquet, O., Navin Lal, T., Weston, J., Sch\"{o}lkopf, B.
``Learning with local and global consistency''.
In: \emph{Advances in Neural Information Processing Systems}, 16, pp.~321--328, 2004.

\bibitem{ZHS04}
Zhou D., Hofmann T., Sch\"{o}lkopf B. ``Semi-supervised learning on directed graphs'',
In {\it Advances in neural information processing systems.}, pp.1633--1640, 2004.

\bibitem{ZB07}
Zhou, D., and Burges, C. J. C.
``Spectral clustering and transductive learning with multiple views''.
In \emph{Proceedings of ICML 2007}, pp.~1159--1166, 2007.

\bibitem{Z05}
Zhu, X. ``Semi-supervised learning literature survey''.
University of Wisconsin-Madison Research Report TR 1530, 2005.

\bibitem{CUDA}
NVIDIA CUDA Programming Guide, available at
{\tt http://docs.nvidia.com/cuda/}

\bibitem{webKB}
Craven, Mark, et al. Learning to extract symbolic knowledge from the World Wide Web. No. CMU-CS-98-122. Carnegie-Mellon Univ Pittsburgh pa School of Computer Science, 1998.
\end{thebibliography}
\end{document}